\documentclass[journal]{IEEEtran}

\ifCLASSINFOpdf

\else

\fi

\usepackage{cite}
\usepackage{amsmath,amssymb,amsfonts}
\usepackage[ruled,vlined]{algorithm2e}
\usepackage{graphicx}
\usepackage{subcaption}
\usepackage{textcomp}
\usepackage{xcolor}
\usepackage{multirow}
\usepackage{comment}
\usepackage{booktabs}
\usepackage{tikz,lipsum,lmodern}
\usepackage[most]{tcolorbox}

\newcommand{\ie}{\textit{i.e.}}
\newcommand{\eg}{\textit{e.g.}}
\newcommand{\CP}[1]{\ignorespaces}

\begin{document}

\title{Probabilistic Approach for Road-Users Detection}

\author{Gledson Melotti, Weihao Lu, Pedro Conde, Dezong Zhao, Alireza Asvadi, Nuno Gonçalves, Cristiano Premebida
	\thanks{Gledson Melotti is with the Federal Institute of Espírito Santo-Brazil, and the ISR-UC at University of Coimbra, Portugal. E-mail: gledson@ifes.edu.br}
	\thanks{W. Lu and D. Zhao are with the Univiversity of Glasgow and the James Watt School of Engineering, UK. E-mail: \{w.lu.1@research.gla, Dezong.Zhao@glasgow \}.ac.uk} 
	\thanks{Alireza Asvadi is with IADYS, France. E-mail: alireza.asvadi@gmail.com}
	\thanks{C. Premebida and P. Conde are with the University of Coimbra and the Institute of Systems and Robotics (ISR), Portugal. E-mail: \{cpremebida, pedro.conde, nunogon\}@isr.uc.pt}
	\thanks{N. Gonçalves is with the University of Coimbra, Institute of Systems and Robotics (ISR), and Portuguese Mint and Official Printing Office, Portugal. E-mail: nunogon@isr.uc.pt}
	\thanks{Manuscript received in 2021.}}


\maketitle

\begin{abstract}
Object detection in autonomous driving applications implies that the detection and tracking of semantic objects are commonly native to urban driving environments, as pedestrians and vehicles. One of the major challenges in state-of-the-art deep-learning based object detection are false positives which occur with overconfident scores. This is highly undesirable in autonomous driving and other critical robotic-perception domains because of safety concerns. This paper proposes an approach to alleviate the problem of overconfident predictions by introducing a novel probabilistic layer to deep object detection networks in testing. The suggested approach avoids the traditional Sigmoid or Softmax prediction layer which often produces overconfident predictions. It is demonstrated that the proposed technique reduces overconfidence in the false positives without degrading the performance on the true positives. The approach is validated on the 2D-KITTI objection detection through the YOLOV4 and SECOND (Lidar-based detector). The proposed approach enables interpretable probabilistic predictions without the requirement of re-training the network and therefore is very practical.
\end{abstract}


\IEEEpeerreviewmaketitle

\section{Introduction}

\begin{figure}[!t]
	\begin{center}
		\begin{minipage}[!t]{0.479\textwidth}
			\includegraphics[width=\textwidth]{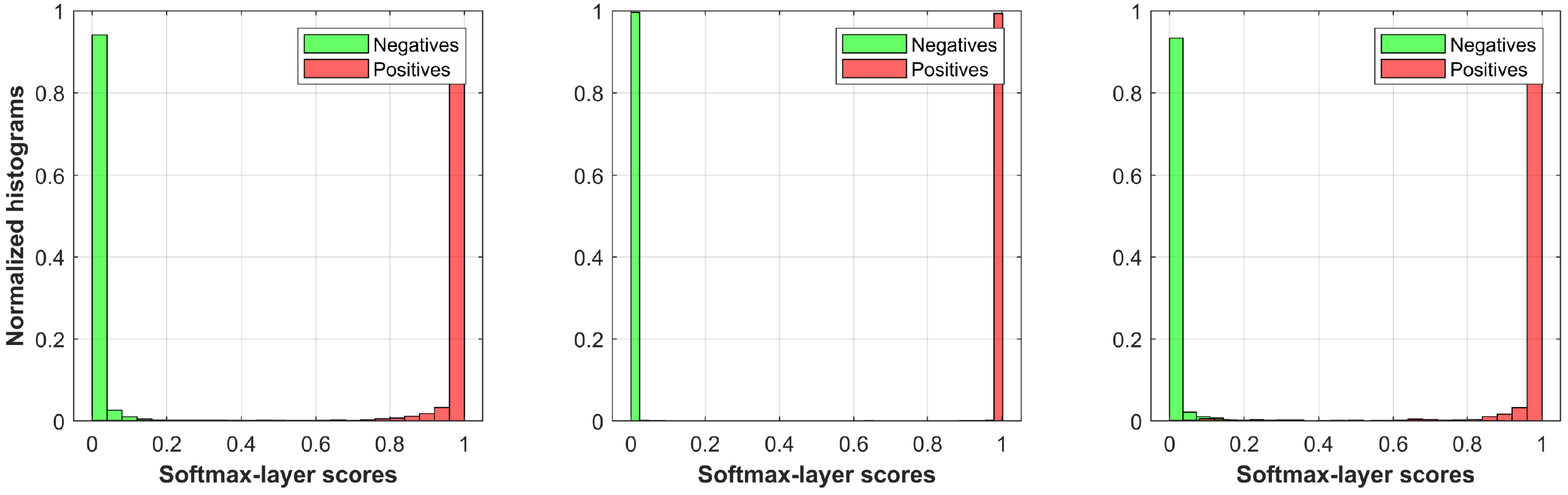}
			\subcaption{Histogram of the Softmax-layer scores.}
			\label{HG_RGB_Softmax}
		\end{minipage}
		\hfill
		\vspace{0.35cm}
		\begin{minipage}[!t]{0.479\textwidth}
			\includegraphics[width=\textwidth]{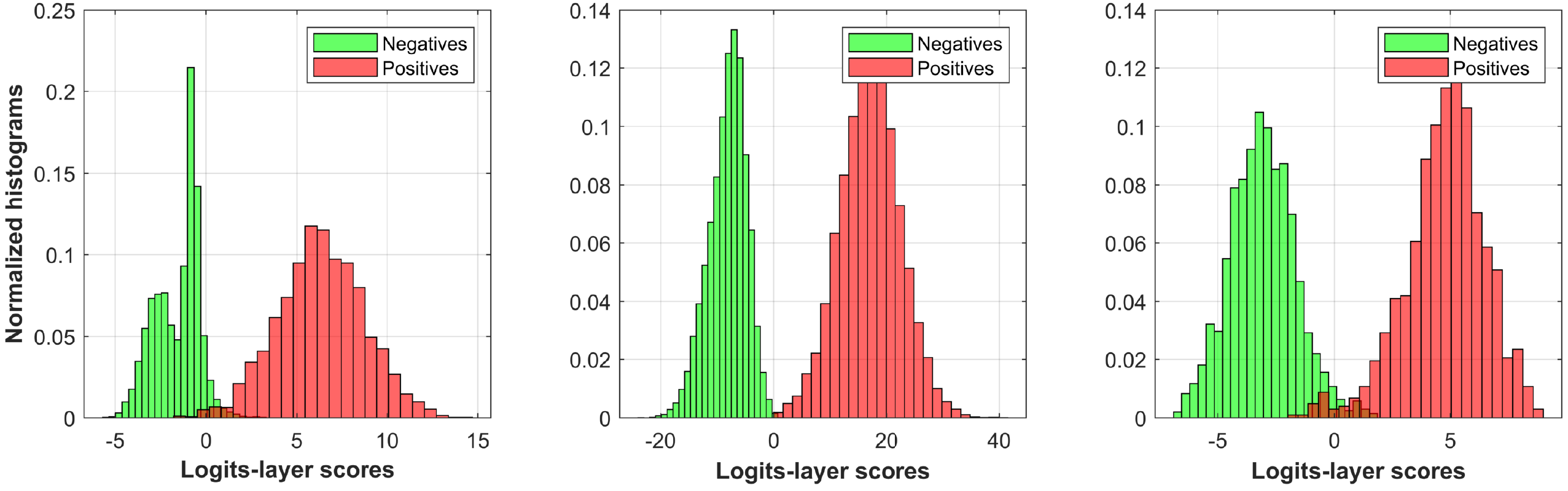}
			\subcaption{Histogram of the logit values.}
			\label{HG_RGB_Logit}
		\end{minipage}
		\caption{In (a) we can see the overconfidence problem regarding the predictions using Softmax for a three classes case (from left to right: pedestrian, car and cyclist). The logit values (\ie, the layer that feeds into Softmax) have been normalized and the corresponding distributions are modelled by a histogram in (b).}
		\label{HG_RGB}
	\end{center}
\end{figure}

\begin{figure}[!t]
	\centering
	\includegraphics[angle=0,scale=0.55]{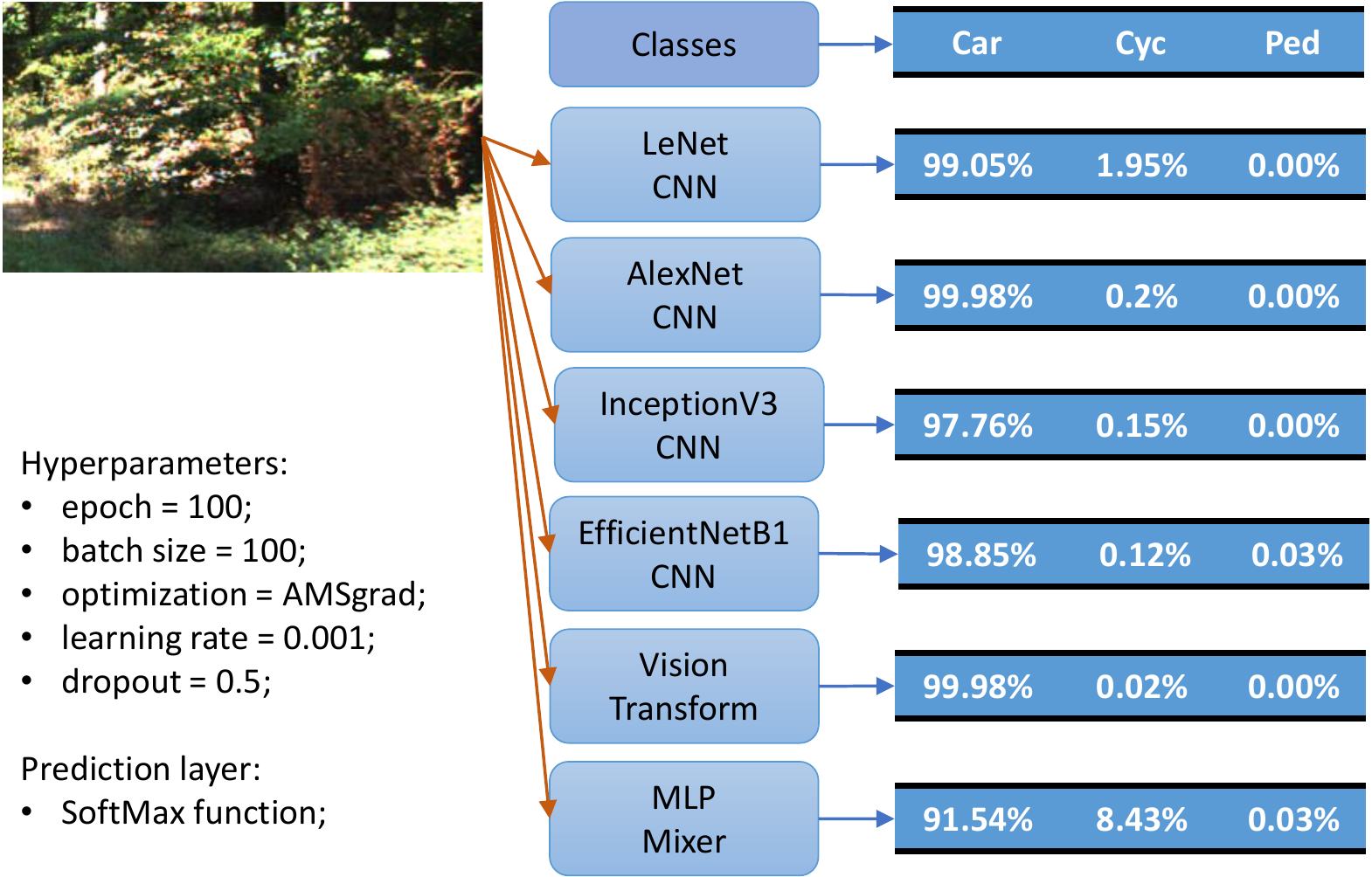}
	\caption{Example of classifying an out-of-(training)-distribution test object. The object has been classified by six different neural networks, and all the models' outputs are overconfident - which may have critical implications.}
	\label{unseen}
\end{figure}

\begin{figure}[!t]
	\centering
	\includegraphics[angle=0,scale=0.355]{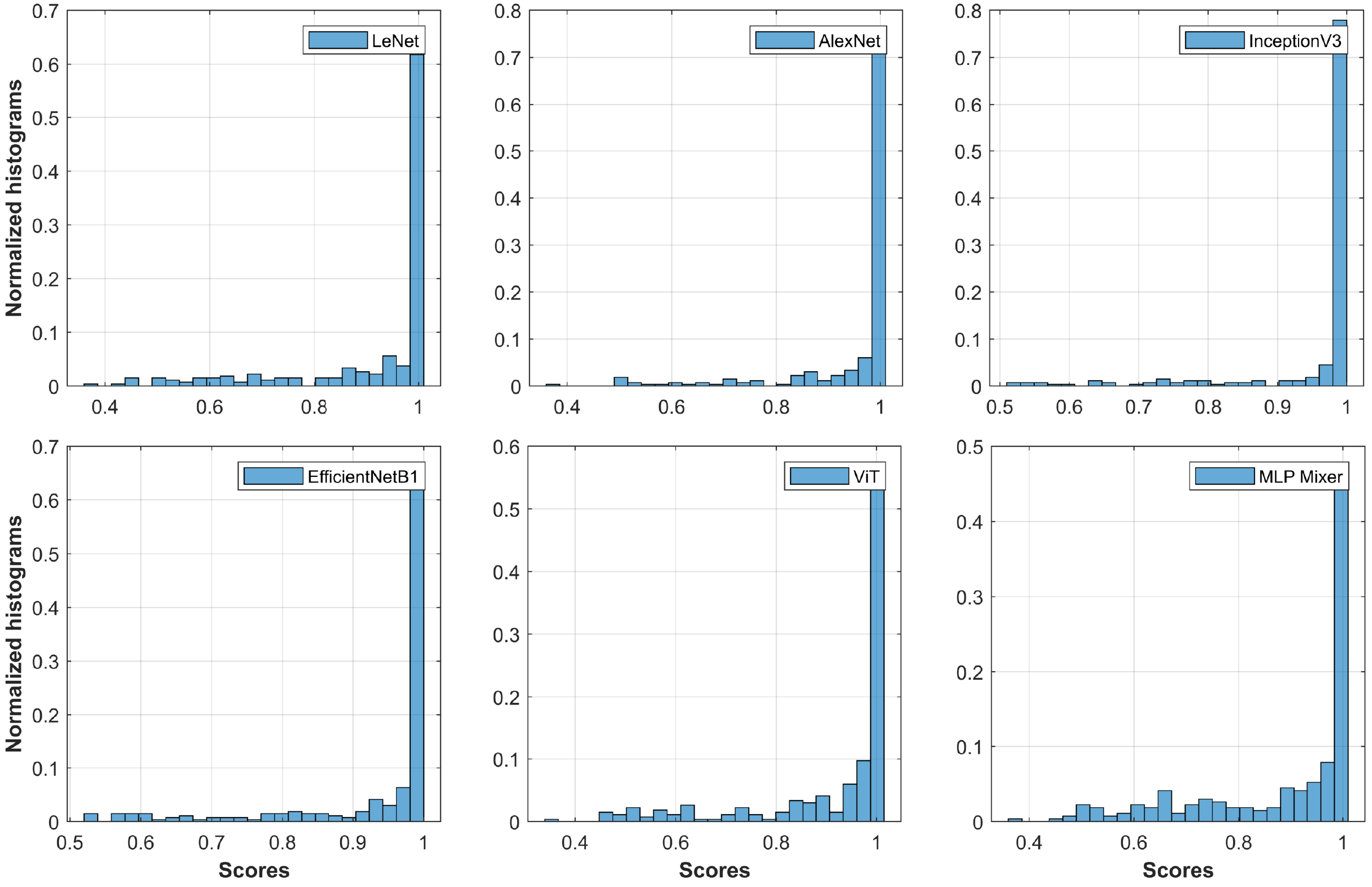}
	\caption{Object classification on out-of-distribution test dataset through six different neural networks, using Softmax as the prediction layer, considering the LeNet~\cite{LeNet}, AlexNet~\cite{Alex2012},  InceptionV3~\cite{Szegedy}, EfficientNetB1~\cite{EfficientNet}, Vision Transformer~\cite{ViT}, and MLP Mixer~\cite{MLP_Mixer} CNNs. The overconfident behavior is notorious.}
	\label{OverConf_Unseen}
\end{figure}

Remarkable advances in computing hardware, sensors and machine learning techniques have contributed significantly to artificial perception for autonomous driving \cite{road,KITTI360,HeQingdong,Janai2019,Shaoshan2017}.
However, even with such progresses, artificial perception in real-world driving still meets grand challenges ~\cite{Claussmann,Janai2019,RobotCarDataset,as}.
Object detection is a key aspect of perception systems and has been gradually dominated by deep learning (DL) approaches. Generally, modern DL methods export the detection confidence as the normalized scores by the Softmax function (SM)~\cite{Su_2018_ECCV} or a single value obtained from the Sigmoid function (SG)~\cite{YOLOV4} without considering the overconfidence or uncertainties in the predictions (see Fig. \ref{HG_RGB}).
\noindent
\raggedbottom
Such a lack of proper uncertainty prediction and the overconfident behaviour are undesired, because objects detected as false positives may have high score values without any level of uncertainty. It can be better understood by an example: consider six deep networks trained to classify three classes of objects namely, car, cyclist, and pedestrian. The detection confidence values for each object have been obtained through a prediction layer, such as the Softmax layer, which then normalizes the values within the interval $[0,1]$. As shown in Table \ref{models_net}, the networks show satisfactory results in terms of F-scores~\cite{Zhang2009,Goutte} on a test set. However, what would happen when an object out of the trained classes is presented to the networks? A clue to answering this question is given by Fig. \ref{unseen}, where an object representing `vegetation' class\footnote{The vegetation class was not considered on the training set.} has been classified with an extremist prediction (\ie, value very close to one, indicating overconfident behaviour) to one of the three trained classes. Ideally, the expected value for that example would be close to 0.3, as the object does not belong to any of the three classes considered in the training. More representative cases of overconfident predictions considering out-of-training distribution examples are shown by histograms in Fig. \ref{OverConf_Unseen}, considering different classes \eg, `person-sitting', `tree', `pole'.

\begin{table}[!t]
	\begin{center}
		\caption{Classification results using F-score metric by deep network models.}
		\begin{tabular}{ccccc}
			\toprule
			Model          & Car     & Cyclist & Pedestrian & Average \\ \hline \hline
			LeNet~\cite{LeNet}         & $99.17$ & $89.08$ & $93.79$    & $94.02$ \\
			AlexNet~\cite{Alex2012}        & $99.42$ & $91.41$ & $96.46$    & $95.75$ \\
			Inception V3~\cite{Szegedy}   & $99.68$ & $95.05$ & $97.67$    & $97.46$ \\
			EfficientNetB1~\cite{EfficientNet} & $99.84$ & $97.43$ & $98.74$    & $98.67$ \\
			ViT~\cite{ViT}            & $99.46$ & $93.56$ & $96.37$    & $96.46$ \\
			MLP Mixer~\cite{MLP_Mixer}     & $98.98$ & $87.47$ & $92.42$    & $92.96$ \\
			\bottomrule
		\end{tabular}
		\label{models_net}
	\end{center}
\end{table}\noindent
\raggedbottom

\noindent
\raggedbottom

\noindent
\raggedbottom

The ability to properly represent the uncertainties of predictions of an object detection system would ensure safer decision-making actions, specially in autonomous driving and robotic systems which may pose threat to people’s lives~\cite{Yarin2017}. 
In the literature, the uncertainties of a deep learning model~\cite{feng21,gledson_eccv,Feng_Di,feng_eccv} can be obtained through the predicted values (calibration techniques) or via the network weights/loss function (regularization techniques)~\cite{patra,Krishnan,Mesquita,Passalis,posch,Feng2019,zouyu2019,oncalibration,GabrielPereyra,gal16,durk,blundell,Kingma,ivalex}. 
However, we will see that calibration and regularization techniques are not immune to the overconfidence problem as well, as detailed in Section \ref{sec_over}. 
An alternative to reduce overconfident predictions, and in some techniques to enable probabilistic interpretation, can be attained by looking at the logit-layer values (\ie, the score-values before the prediction layer, or activation functions) \cite{feng21,gledson_eccv,Feng_Di,feng_eccv} - as illustrated in Fig. \ref{HG_RGB_Logit} which presents a more tractable distribution than the distribution out of the Softmax prediction layer.

In this context, this paper presents a new methodology to reduce overconfident predictions in deep object detection networks without interfering in the cost function and/or re-training the network. Furthermore, this paper shows that calibration techniques (such as temperature scaling and Monte Carlo Dropout, as well as confidence penalty, and Bayesian neural networks) may provide overconfidence results.

In summary, the contributions are:
\begin{itemize}
	\item An investigation of the predicted values using distributions from the logit-layer data;
	\item An efficient way to obtain proper probabilistic inference via Maximum Likelihood (\textit{ML}) and Maximum a-Posteriori (\textit{MAP}) formulations;
	\item Detailed comparisons between the \textit{ML/MAP} against the Sigmoid layer, considering true and false positive predictions by YOLOV4 and SECOND, with respect to overconfidence results;
	\item Comprehensive results showing that the traditional prediction layers can induce erroneous decision-making in deep object detection networks.
\end{itemize}

\section{Related Work on Overconfident Predictions}
\label{sec_over}

Generally, the formulations that acts directly on the predicted scores to reduce overconfident predictions of learning models are considered as post-processing (or \textit{post-hoc}) calibration techniques~\cite{Jiacheng,Frenkel,oncalibration,isotonicregression,plattscaling}. On the other hand, the problem of overconfident predictions in deep models, can also be addressed with regularization techniques (formulations that interfere with the learning procedure of the model, to improve the generalization ability)~\cite{Weizh_pmlre,Rafael,Szegedy}, Bayesian models (that leverage approximate Bayesian inference instead of classical point estimation in neural networks)~\cite{Kendall2017,Kingma,Bishop}, or even augmentation methods \cite{Conde_2022_BMVC}, that produce better-calibrated models. Well-calibrated models are expected to provide accurate predictions when they are right about object detection and, conversely, provide high uncertainty when they are inaccurate about a detection. However, such techniques to reduce or mitigate overconfidence are still to be improved~\cite{Krishnan}. Actually, recent studies have shown overconfident predictions as unsolved problems in the field of deep learning~\cite{Krishnan,kristiadi2020,thulasidasan2019,bulatov2015,raudys2003}. Consequently, several probabilistic methods have been proposed as an alternative to reduce overconfident predictions, as well as to capture uncertainties in deep neural network models~\cite{feng21,gledson_eccv,Feng_Di,feng_eccv,Mesquita,Passalis,posch,Feng2019,zouyu2019,Feng2018,flipout2018,Kendall2017,Balaji,GabrielPereyra,gal16,durk,blundell,Kingma,ivalex}.

The following subsections present more details about the most common and recent calibration techniques (like temperature scaling~\cite{oncalibration}), some regularization techniques (penalization of overconfident output distributions~\cite{posch,zouyu2019,flipout2018,Balaji,GabrielPereyra}, label smoothing~\cite{lukasik20a}) and some forms of approximate Bayesian inference (like variational inference \cite{Kingma} and Monte Carlo Dropout~\cite{ConcreteB,gal16}). Additionally, we would discuss the disadvantages of the mentioned techniques when predicting objects belonging to out-of-training-distribution data (which may be critical in autonomous driving and robotics).

\subsection{Softmax and Sigmoid Prediction Layers}

The Softmax function, a generalization of the Sigmoid function for the multiclass case, is currently one of the most commonly employed functions to act as the prediction layer in deep networks. In part, this is explained by the fact that such function increases the weights of the correct classes in an exponential way, strongly interfering in the updating of the weights, and thus may guarantee a better result in terms of classification performance. However, such behaviour may lead to overfitting, since the model becomes overconfident on the training data~\cite{Relaxed}. Additionally, the Softmax function does not provide any reliable confidence measurements for the predicted values~\cite{corbi,oncalibration,HendrycksG17}. Also, it is possible to find in the literature works where the Softmax's outputs are considered actual likelihood values~\cite{Jiacheng,Bingyuan,Youngbum,Wang_2021_ICCV} (perhaps because they sum up to one) which tends to give an erroneous  probabilistic interpretation about the results.

The Softmax, as well as the Sigmoid function, are sensitive to adversarial attacks. The studies that back this claim consider adversarial perturbations applied to the Softmax and Sigmoid prediction layer, generating possible underfitting problems on the weights~\cite{goodfellow2015,ChrSzegedy}. Additionally to the fact that Softmax and Sigmoid functions are prone to provide poorly calibrated scores and being sensitive to adversarial attacks, such functions also seem to be inadequate to cope with out-of-distribution objects in the test phase (\textit{e.g.}, during the evaluation time the trained network can be faced with objects that do not fit to any of the training classes) as demonstrated experimentally in~\cite{gledson_eccv,LeaConf,LiangLS18,HendrycksG17,Balaji,Gal2016}.

\subsection{Post-processing Calibration Techniques}
Among the various existing techniques to reduce overconfident predictions, post-processing calibration techniques present the advantage of being easily applied to pre-trained models. For example, temperature scaling has demonstrated interesting characteristics because it is simple and, in some cases, efficient~\cite{oncalibration}. 

The value of temperature scaling ($TS$) is obtained by minimizing the negative log likelihood (NLL) on the validation set. All the values of the logit vector (before the prediction layer) are multiplied by a scalar parameter $\frac{1}{TS}$, with $TS>0$. \CP{, during the prediction at test time.} Simply, the temperature scaling parameter can be included in the Softmax prediction layer (SM)
\begin{align}
	\label{eq_sm}
	SM(\hat{z_j}) = \frac{e^{(\hat{z_j}/TS)}}{{\displaystyle \sum_{k=1}^K e^{(\hat{z_k}/TS)}}},
\end{align}
where $k\in \{1, \ldots, K\}$, $K$ is the number of classes, $\hat{z_j}$ is the output of the predicted logit layer \ie, predict score value of the object $j$. 

\subsection{Regularization Techniques}

Different from the post-processing techniques, regularization techniques such as label smoothing and confidence penalty act during the training process, on the updates of the weights according to the cost function~\cite{Rafael,lukasik20a,GabrielPereyra,Szegedy}. 

For classification problems, defining $\mathbf{X}=\{\mathbf{x_1}, \ldots, \mathbf{x_j} \}$ as input data, and $\mathbf{Y}=\{\mathbf{y_1}, \ldots, \mathbf{y_j} \}$ as output data obtains the dataset $D = \{\mathbf{x_j},\mathbf{y_j}\}_{j=1}^{N_{ts}}$, where $N_{ts}$ is training set size, $\mathbf{x_j} \in R^n$, and $\mathbf{y_j}\in \{1, \ldots, K\}$ with $K$ classes, the loss function considering the true label as one-hot encoding vector is defined by
\begin{align}
	\mathcal{L} = - \frac{1}{N_{ts}}\sum_j^{N_{ts}} p(\mathbf{y_j}|\mathbf{x_j})\mbox{log}(p(\mathbf{\hat{y_j}}|\mathbf{x_j})), \label{eq_ce_7}
\end{align}
where $p(\mathbf{y_j}|\mathbf{x_j})$ is the distribution of the true label (ground-truth) given the data, $\mathbf{\hat{y_j}}$ is the predicted value for the input $\mathbf{x_j}$, and $p(\mathbf{\hat{y_j}}|\mathbf{x_j})$ is the predicted labels distribution.
The expression of the confidence penalty (\ref{eq_me_1}) includes a weighting term in the cost function given in (\ref{eq_ce_7}). The additional term is the Entropy of the predicted values, and $\beta$ is the parameter that controls the confidence penalty~\cite{GabrielPereyra}
\begin{align}
	\mathcal{L} =-\frac{1}{N_{ts}}\sum_{j}^{N_{ts}} [p(\mathbf{y_j}|\mathbf{x_j}) \mbox{log}(p(\mathbf{\hat{y_j}}|\mathbf{x_j})) \nonumber\\
	-\beta p(\mathbf{\hat{y_j}}|\mathbf{x_j})\mbox{log}(p(\mathbf{\hat{y_j}}|\mathbf{x_j}))].
	\label{eq_me_1}
\end{align}

Unlike confidence penalty, the label smoothing technique does not interfere with the mathematical formulation of the cost function, making the model less certain about the provided predictions. In fact, label smoothing modifies the values of the one-hot encoding vector, as defined in (\ref{eq_target})~\cite{Szegedy}
\begin{align}
	\mathbf{y_{new_{j,k}}} = (1-\epsilon)\mathbf{y_{j,k}}+\frac{\epsilon}{K},
	\label{eq_target}
\end{align}
where $\mathbf{y_{j,k}}$ is the object $j$ in the class $k$, $\mathbf{y_{new_{j,k}}}$ is the new label value, $\epsilon$ is the smoothing parameter arbitrarily defined, and $K$ is the number of classes. Label smoothing reduces the difference between the values of the labels of the correct class against the values of the other classes, interfering in the updating of the weights of the network. Not using the label smoothing technique can cause two problems, according to~\cite{Szegedy}: ``First, it may result in over-fitting: if the model learns to assign full probability to the groundtruth label for each training example, it is not guaranteed to generalize. Second, it encourages the differences between the largest logit and all others to become large, and this, combined with the bounded gradient $\cfrac{\partial l}{\partial z_k}$, reduces the ability of the model to adapt. Intuitively, this happens because the model becomes too confident about its predictions''.

\subsection{Bayesian Neural Networks}
Bayesian Neural Networks are modelled using approximate Bayesian inference (\ref{bayes_1}) to assign probabilities to events, and thus capturing uncertainties in a model's predictions \cite{Kendall2017,Balaji,Bishop}, by considering the network weights as a probability distribution parameter(s) instead of a `deterministic' value (like in traditional deep neural networks). The posterior probability of the weights given the input and the target/class data can be expressed by~\cite{Kendall2017,gal16}
\begin{align}
	\label{bayes_1}
	p(\mathbf{W|X,Y})=\cfrac{p(\mathbf{Y|X,W})p(\mathbf{W})}{p(\mathbf{Y|X})},
\end{align}
where $\mathbf{W}=\{\mathbf{w_1}, \ldots, \mathbf{w_i} \}$ denotes the weights matrix, $\mathbf{X}$ is input data, $\mathbf{Y}$ is output data, $p(\mathbf{W})$ is the prior distribution, which expresses the uncertainty before any data observed~\cite{GoodBengCour,Bishop}, and $p(\mathbf{Y|X,W})$ is the class conditional density (likelihood function). The $p(\mathbf{Y|X})\neq 0$ acts as a scaling factor for $p(\mathbf{W|X,Y})$, and it can be expressed as $\int p(\mathbf{Y|X,W})p(\mathbf{W})d\mathbf{W}$ that can often be determined by the law of the total probability~\cite{Bishop}. For example, considering a discrete case\footnote{Probability formulations for continuous cases are represented by lowercase letters, while for discrete cases they are represented by uppercase letters.}, $P(\mathbf{Y|X})$ can be computed per parameter $\mathbf{w_i}$ \ie, $\sum P(\mathbf{Y|X},\mathbf{w_i})P(\mathbf{w_i})$. 

The calculation of the posterior $p(\mathbf{W|X,Y})$ may not be trivial because the density function $p(\mathbf{Y|X})$ can assume a complex form (whereas the prior can be specified from some previous knowledge and the likelihood conceivably obtained from the data). For this reason, in complex models - like deep neural networks - the posterior becomes intractable. Thus, a possible solution is to perform an approximation by means of variational inference~\cite{Kingma,flipout2018,ConcreteB,Molchanov,GoodBengCour,blundell,durk,Bishop}. Nonetheless, variational inference still presents some challenges in terms of computational complexity, specially when dealing with large models and large quantities of data.

A computationally more efficient (and therefore popular) method of approximate Bayesian inference is the Monte Carlo Dropout formulation,~\cite{gal16, ConcreteB}, that leverages \textit{dropout} ~\cite{srivastava2014dropout} (commonly used as a regularization technique) at test time, to capture the model uncertainty. Dropout~\cite{srivastava2014dropout} is a stochastic technique~\cite{Gal2016}, which might potentially be included in the neural network, contributing to avoid overfitting. It is usually used during training, and therefore it can be questioned: what does occur when the dropout is used during testing? The predicted values will not be deterministic \ie, the values depend on which connections between the neurons will be randomly chosen during the prediction stage. In fact, the same test sample forwarded several times in the network can have different predicted values. In \cite{gal16}, the authors show that applying dropout (at inference) before every weight layer of a deterministic deep neural network is equivalent to an approximation of a probabilistic deep Gaussian process.

\begin{figure}[!t]
	\begin{center}
		\begin{minipage}[!t]{0.48\textwidth}
			\includegraphics[width=\textwidth]{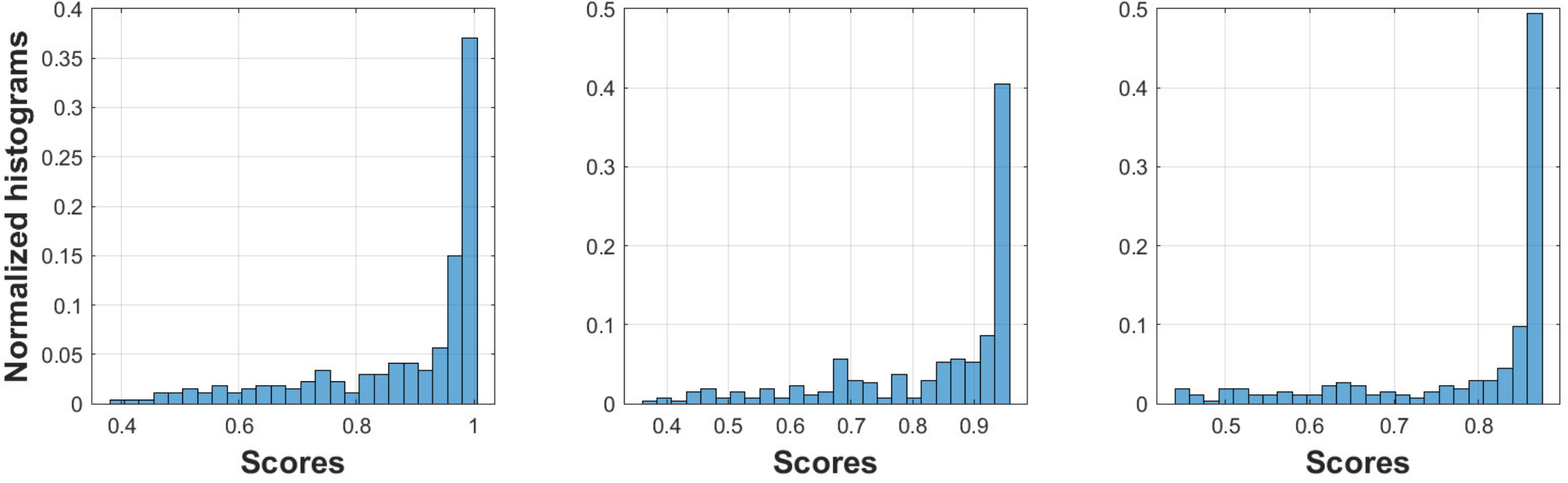}
			\subcaption{From left to right: temperature scaling~\cite{oncalibration}, confidence penalty and label smoothing~\cite{GabrielPereyra}.}
			\label{UnseenTS}
		\end{minipage}
		\hfill
		\vspace{0.35cm}
		\begin{minipage}[!t]{0.48\textwidth}
			\includegraphics[width=\textwidth]{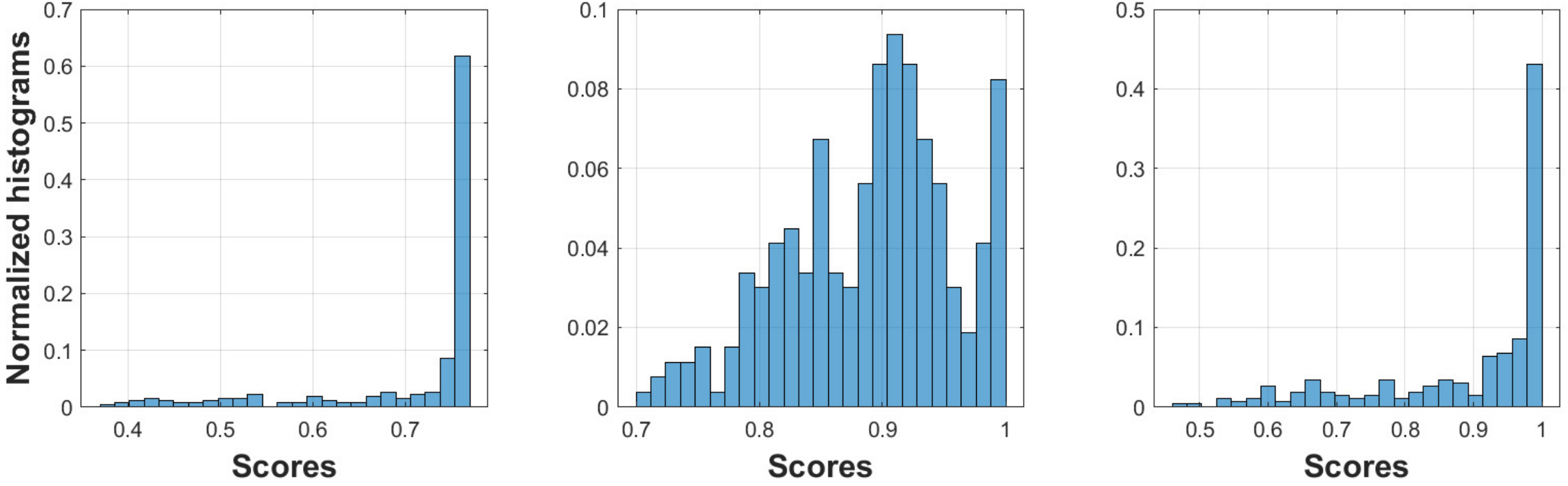}
			\subcaption{From left to right: confidence penalty with label smoothing~\cite{GabrielPereyra}, Monte Carlo Dropout~\cite{ConcreteB}, and Bayesian neural network.}
			\label{UnseenPC}
		\end{minipage}
		\caption{Object classification on out-of-(training)-distribution test dataset using calibration and regularization techniques in an InceptionV3 CNN model.}
		\label{RGSLBNN_Unseen}
	\end{center}
\end{figure}\noindent
\raggedbottom

\subsection{Discussion on the State of the Art}

Temperature scaling, confidence penalty, and label smoothing techniques aim to reduce the overconfidence problem when making predictions using relatively simple formulations. Temperature scaling also enables, as an advantage, the possibility of being applied without the need to re-train the network. The disadvantage of these techniques is the inability to directly provide an uncertainty interval regarding the detected objects subjected to the trained classes. Monte Carlo Dropout and Bayesian neural networks, on the other hand, provide uncertainties measures \ie, the mean and variance associated with each confidence value, but with relatively higher computational cost.

Figure \ref{RGSLBNN_Unseen} shows the performance of some of the previously mentioned techniques by considering out-of-distribution test objects (person sitting, tree, pole/stem). The networks were trained from scratch to classify objects belonging to the categories $\{$car, cyclist, pedestrian$\}$, considering $\epsilon=0.2$ in (\ref{eq_target}) for label smoothing, $\beta=0.3$ in (\ref{eq_me_1}) for the confidence penalty, $TS=1.82$ in (\ref{eq_sm}) for temperature scaling, and for Monte Carlo Dropout the test sample was forwarded $300$ times through the network. In the case of the Bayesian neural network, the classification experiments were conducted using the Tensorflow toolbox. Note that most of the objects in this controlled experiment have been classified with overconfidence.

The overconfidence problem in deep models can be detrimental to draw a firm conclusion regarding safety, particularly because it is not possible to foresee all kinds of objects that can appear, for example, within a perception system's FOV of an autonomous vehicle operating in a real-world (uncontrolled) environment. However, it can be partially concluded that the behavior shown in Fig. \ref{RGSLBNN_Unseen} makes it very difficult to interpret the model's confidence in a proper way.

\section{Probabilistic Inference For Object Detection}
This section presents a formulation to reduce overconfident
predictions on existing deep object-detectors, including non-parametric and parametric modeling to represent the likelihood and the priors. The proposed approach relies on a Maximum Likelihood (ML) and Maximum a-Posteriori (MAP) function-layers, based on the Bayes’ rule, to replace Softmax or Sigmoid functions depending on the object detector.

\subsection{ML and MAP Layers}
\label{subsec.ML_MAP}

The formulation behind the Bayesian inference for the proposed ML and MAP layers is built up from the logit outputs/scores (denoted by $\mathbf{x}$) and the random variables $\mathbf{C}$ and $\mathbf{W}$ \ie, the class-labels and the network weight respectively. The decision layers will then output a posterior $P(\mathbf{C|x,W})$ that is proportional to the class-conditional density (\ie, likelihood) $p(\mathbf{x|C,W})$ and the priors $P(\mathbf{C})$, where $\mathbf{C}=\{c_1,\ldots,c_N\}$ and $\mathbf{x}=\{x_1,\ldots,x_N\}$, with $x_i$ corresponding to the logit value for the class $c_i$. Thus, the Bayes' rule may simply be given by (\ref{bayes0}), considering that the weights were the result of a learning process in order to explain the data~\cite{GoodBengCour} and are assumed to be constant after the training,
\begin{align}
	P(\mathbf{C}|\mathbf{x})=\cfrac{p(\mathbf{x}|\mathbf{C})P(\mathbf{C})}{p(\mathbf{x})}.
	\label{bayes0}
\end{align}

The law of total probability \cite{Bishop,Papoulis} allows (\ref{bayes0}) to be rewritten using the \textit{per-class} discrete formulation,
\begin{align}
	P(c_i|\mathbf{x}) = \cfrac{P(\mathbf{x}|c_i)P(c_i)}{{\sum\limits_{i=1}^{K}P(\mathbf{x}|c_i)P(c_i)}},
	\label{bayes1}
\end{align}
where $K$ is the number of classes.

Inference can then be made on the test set regarding $\mathbf{C}$ given the dependence with $\mathbf{x}$ \ie, the value of the posterior probability (\ref{bayes1}) of $\mathbf{C}$ is determined after observing the value of $\mathbf{x}$. Once we have specified the likelihood distribution $p(\mathbf{x|C})$, and the priors, the proposed ML/MAP prediction layers can be used to replace a Softmax or a Sigmoid function in order to output the object classification scores in a probabilistic way\CP{ output, taking into account the object classification scores in the perception system as being the probabilities over a categorical distribution}. Thus, the Maximum Likelihood (\textit{ML}) and Maximum a-Posteriori (\textit{MAP}) functions can be defined as prediction layers at the testing time, and they are expressed by 
\begin{align}
	ML &= arg \max_{i} \cfrac{(P(\mathbf{\textbf{x}}|c_i)+\lambda)}{\sum\limits_{i=1}^{K}(P(\mathbf{\textbf{x}}|c_i)+\lambda)},\label{bayes2}\\
	MAP &= arg \max_{i} \cfrac{(P(\mathbf{\textbf{x}}|c_i)P(c_i)+\lambda)}{\sum\limits_{i=1}^{K}(P(\mathbf{\textbf{x}}|c_i)P(c_i)+\lambda)},\label{bayes3}
\end{align}
where $\lambda$ is an additive smoothing parameter to avoid the ``zero'' probability issue \cite{AdditiveS,SmoTec,Lidstone}, to indirectly mitigate the overconfidence problem, and at the same time incorporate some unpredictable level of uncertainty in the final prediction\CP{(mainly motivated by out-of-distribution objects)}. The parameter $\lambda$ is not too high or too small, and does not depend on any specific prior information, but its value has to preserve \CP{. The respective parameter can be determined empirically by observing which value preserves} the original distribution `shape' without degrading the final result.

Notice that, although the Bayesian formulation takes distributions into account, $ML$ and $MAP$ layers compute a single estimate rather than a distribution.

\subsection{Estimating the Likelihood and Prior Probability}
\label{subsec.pdf_prior}

\begin{figure*}[!t]
	\centering
	\includegraphics[scale=0.545]{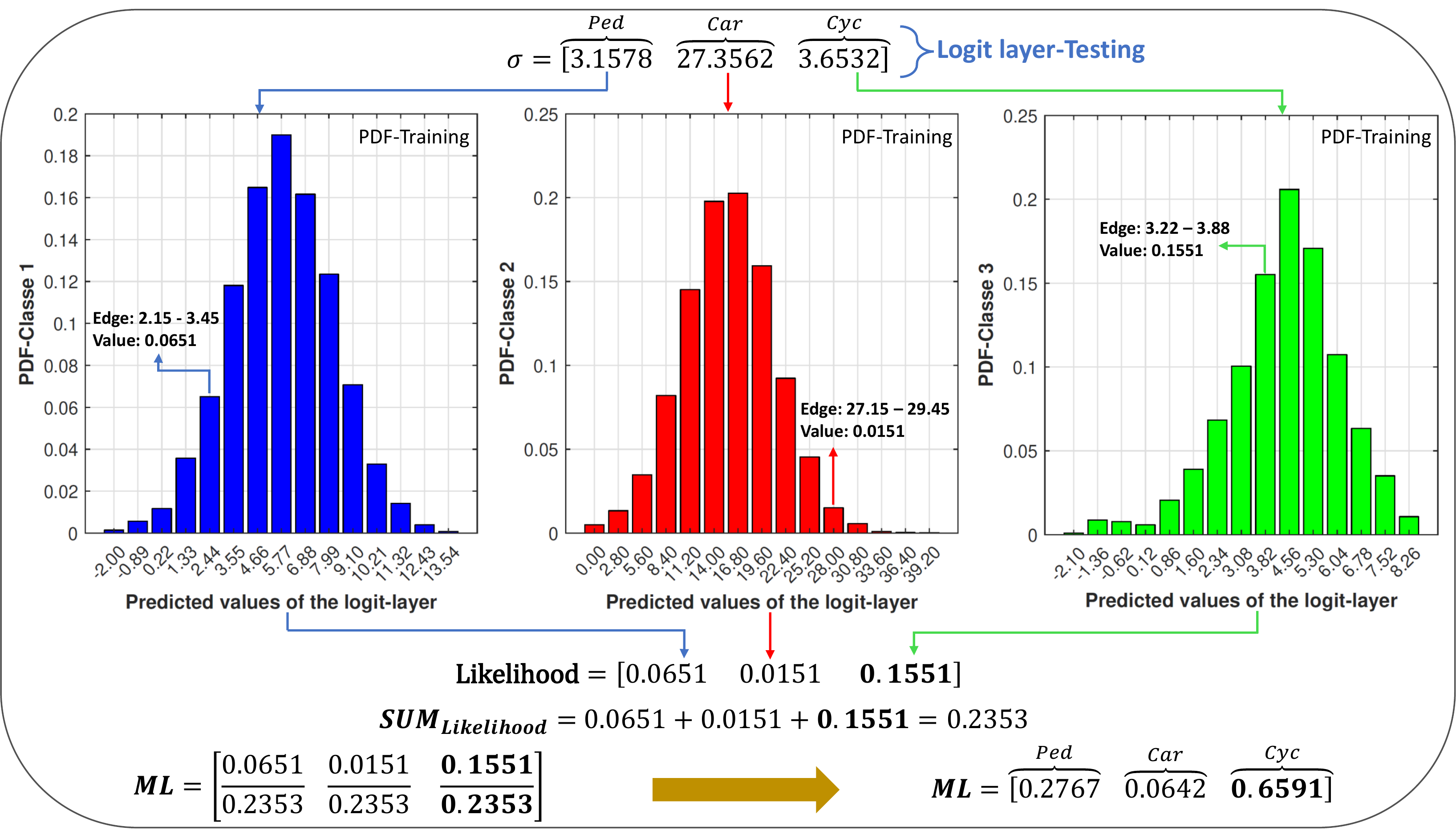}
	\caption{Getting the probability values from normalized-histograms used to model the distributions of the logits on the training set. \CP{from the histograms by class using the logit-layer values in the training set}}
	\label{fig_likelihood}
\end{figure*}

The non-parametric probabilistic density distribution chosen here to obtain the likelihood function comes from normalized histograms\footnote{The importance of normalizing the histogram is to ensure that the sum of the probabilities is one.} of the logit-layer's scores for each class on the training dataset, as shown in Fig. \ref{fig_likelihood}

During the testing phase (\ie, on the test set), the logit-layer score per example (or object) will then be matched to the per-class histogram, as illustrated in Fig. \ref{fig_likelihood}.
 \noindent
\raggedbottom

\begin{figure}[!t]
	\centering
	\includegraphics[scale=0.695]{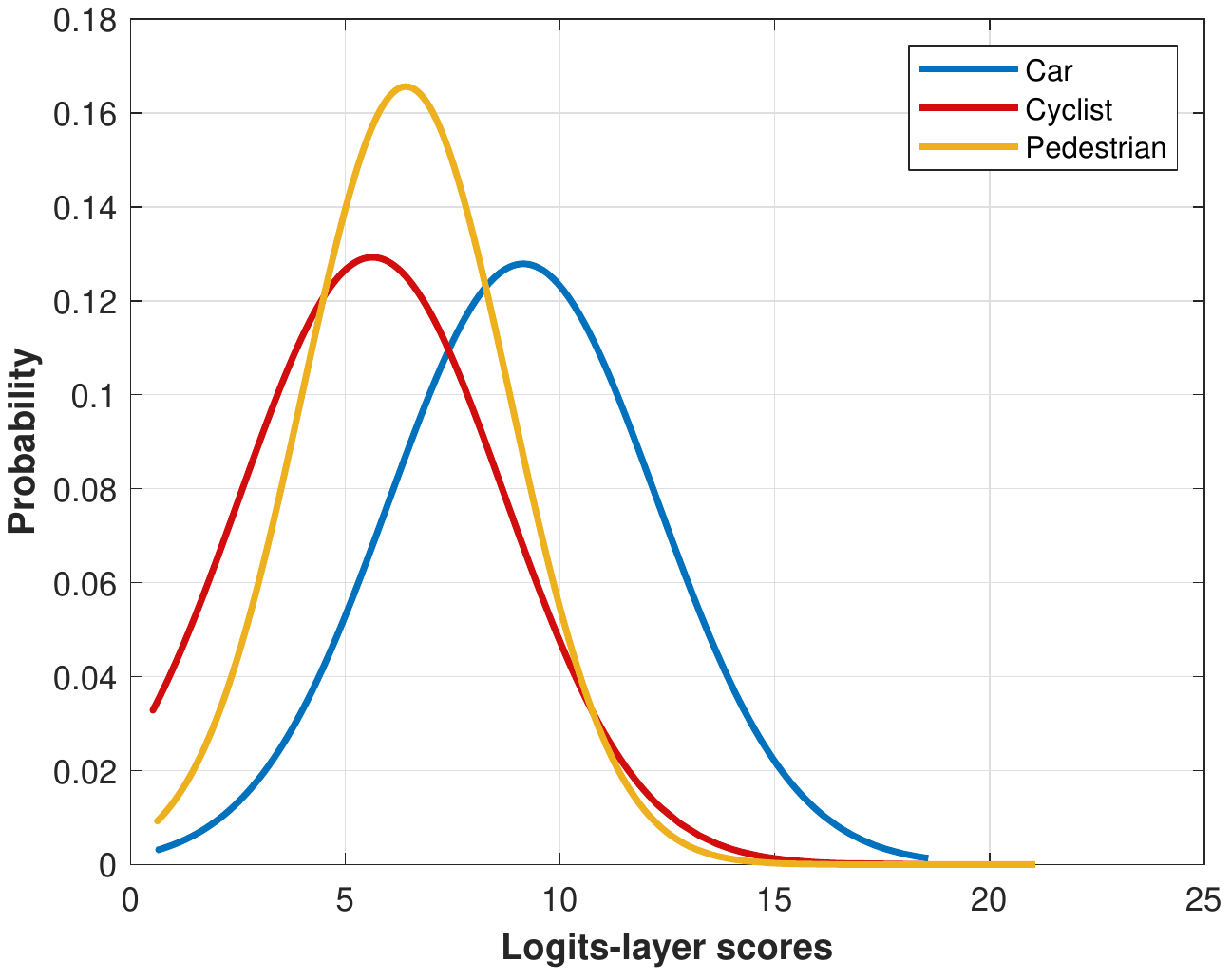}
	\caption{Gaussian distributions to estimate the prior probabilities for the three training classes (car, cyclist and pedestrian).}
	\label{fig_prior}
\end{figure}

Unlike the likelihood function estimation, the prior probability distribution has been modelled by a Normal. Thus, the parametric estimation depends on the mean and the variance obtained from the logit scores as well (this time it is a continuous pdf as shown in Fig. \ref{fig_prior}). Therefore, the prior is $P(c_i)\sim \mathcal{N}(\mathbf{x}|\mu,\,\sigma^{2})$ with mean $\mu$ and variance $\sigma^{2}$ computed per class.
 \noindent
\raggedbottom

The purpose of considering a discrete (normalized histogram) and a continuous pdf to model the likelihood and the \textit{a-prior} probability respectively, is motivated from the perspective of complementary information that can be extracted from the same data. 

Algorithm $1$ summarizes the steps of the proposed methodology to computes \textit{ML} and \textit{MAP} layers scores of each class from the logit-layer values.Note that some detection models consider the objectness score ($OS$) parameter (parameter obtained during training), according to YOLOV4. $OS$ is a parameter which defines whether a region in the image (grid) contains an object or not. For each grid in the image, the network provides a set of bounding-boxes, having each bounding-box an objectness score and a classification score. From an objectness threshold, the network defines which is the best bounding box that represents a given object. In other words, $OS$ is used to evaluate which bounding box centered on a grid best represents the detected object~\cite{YOLOV4}. By multiplying $OS$ with the classification score, the resulting \CP{in}is the confidence level of the detected object. Thus, in the formulation of YOLOV4, the final process of defining an object's class is to multiply the objectness score with the classification score. Therefore, the proposed methodology maintains the same way of classifying an object according to the detection algorithm being analyzed. In other words, in the case of YOLOV4, the proposed methodology replaces the classification scores obtained by the Sigmoid function by the scores from the ML and MAP layers \ie, multiplying the \textit{ML} and \textit{MAP} scores by the objectness scores.
\begin{figure}[!t]
	\begin{tcolorbox}[colback=black!5!white,colframe=black!75!black,title = Algorithm 1: \textit{ML} and \textit{MAP} Layers]
		\textbf{Input:}
		\begin{itemize}
			\item Densities (normalized histogram and Gaussian distribution on the training set - logit-layer values, Fig. \ref{fig_likelihood});
			\item Logit-layer values on the test set ($Test$);
			\item Additive smoothing ($\lambda$);
			\item Number of classes ($K$).
		\end{itemize}
		
		\textbf{Output:}
		\begin{itemize}
			\item Maximum Likelihood (\textit{ML});
			\item Maximum a-Posteriori (\textit{MAP}).
		\end{itemize}
		
		\textbf{Normalized frequency histograms:}\\
		$hc \gets histogram(Train($K$))$;\\
		\textbf{Edge values of each bin of each histogram:}\\
		$BinLow \gets BinEdgesLow(hc)$;\\
		$BinHigh \gets BinEdgesHigh(hc)$;\\
		\textbf{Frequency values of each of the histograms:}\\
		$V \gets Values(hc)$\\
		\textbf{Getting the likelihood:}\\
		$P(\mathbf{x|C}) \gets zeros(size(Test),K)$;\\
		\For{$k \gets 1:size(Test)$}{
			\For{$cl \gets 1:K$}{
				\For{$i \gets 1:size(BinValues)$}{
					\If{$(BinLow(cl,i) \leqslant Test(k,cl))\\
						\, \& \, (Test(k,cl) < BinHigh(cl,i))$}
					{
						$P(\mathbf{x}|C)(k,cl) \gets V(cl,i)$\;
					}
					end}
				end}
			end}
		\textbf{Getting the Prior:}\\
		$P(\mathbf{C}) \gets \mathcal{N}(Test|[\mu_{Train},\,\sigma^{2}_{Train}])$;
		
		\textbf{Calculating the \textit{ML}  and \textit{MAP}:}\\
		$ML \gets P(\mathbf{x|C}) + \lambda$;\\ 
		$ML \gets (ML/\mbox{sum}(ML))*\mbox{ObjectnessScore}$;\\
		$MAP \gets P(\mathbf{x|C})P(\mathbf{C}) + \lambda$;\\ 
		$MAP \gets (MAP/\mbox{sum}(MAP))*\mbox{ObjectnessScore}$;
	\end{tcolorbox}
\end{figure}\noindent
\raggedbottom

\section{OBJECT DETECTION}

\begin{figure*}[!t]
	\centering
	\includegraphics[scale=0.54]{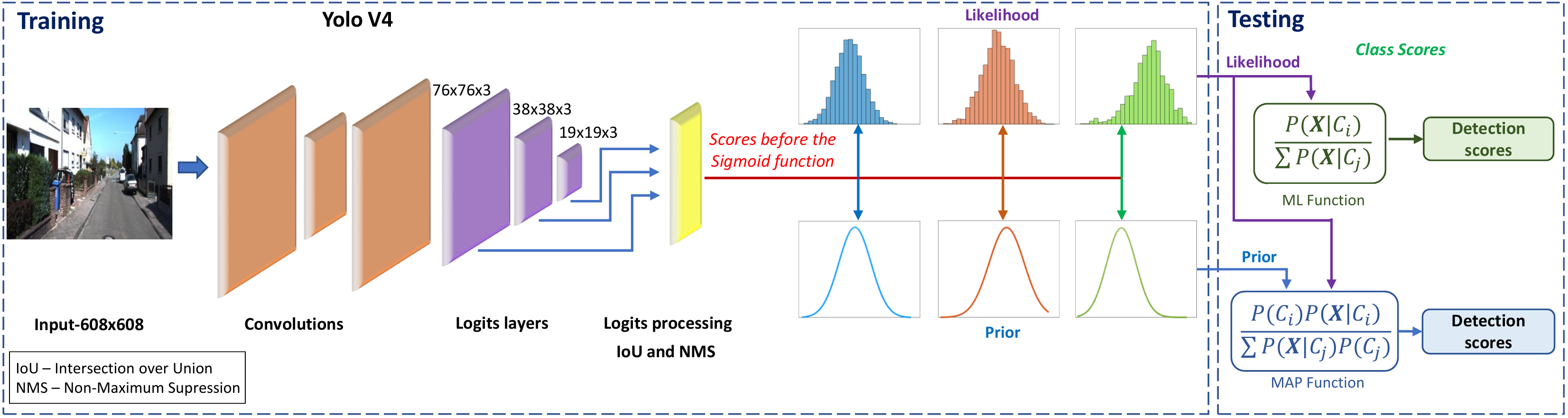}	
	\caption{YOLOV4 representation with logit and Sigmoid (\textit{SG}) layers, Maximum Likelihood (\textit{ML}) and Maximum a-Posterior (\textit{MAP}) functions. After training, the predicted values from the Sigmoid Layer were replaced by the scores from \textit{ML} and \textit{MAP} functions. Notice that the YOLOV4 was not trained or re-trained with the \textit{ML/MAP} functions.}
	\label{YOLOV4}
\end{figure*} \noindent
\raggedbottom

Currently, the state of the art in pattern recognition for autonomous driving and robotics is closely related to object detection using deep models, which has become one of the most important areas of computer vision (including LiDAR-based systems). The primary purpose of a detector is to estimate the object's position, size and class/category. A $2D$ detector estimates bounding boxes considering the coordinates of the center, width and height of the objects' hypothesis. Additionally, detectors estimate the classification score and predicted class. In plain words, the recent detectors rely on a series of steps to define the bounding boxes and the classification scores depending on comparisons across thresholds between predicted output and ground-truth (training stage), as well as objectness score threshold, intersection over union (IoU), non-max suppression (NMS), and class threshold.

Among the various detection models, we have chosen the YOLOV4~\cite{YOLOV4}, published in 2020, which at the time has reached the state of the art performance on the COCO dataset, while achieving shot inference time. The structure of YOLOV4 and the proposed methodology is illustrated in Fig. \ref{YOLOV4}.

The advantages of YOLOV4, over previous versions and other existing object detection algorithms, are that YOLOV4 tries to avoid overconfident results by using data augmentation (CutMix and Mosaic), class label smoothing, and dropout in the convolution layers (DropBlock regularization), which then influence the classifier accuracy. Also, unlike many object detection algorithms, YOLOV4 uses the Mish activation function instead of the traditional functions (\eg, ReLU, ELU, SeLU, PReLU, Swich). Additionally, the cost function of YOLOV4 incorporates overlap area, central point distance and aspect ratio~\cite{CIOU}, as well as cosine annealing scheduler (learning rate)~\cite{cosinerate}, a modified cross-iteration batch normalization~\cite{CBN}, self-adversarial training~\cite{YOLOV4}. Finally, the Sigmoid function is employed to get the final bounding boxes and the respective classification scores.

\begin{figure}[!t]
	\begin{center}
		\begin{minipage}[!t]{0.48\textwidth}
			\includegraphics[width=\textwidth]{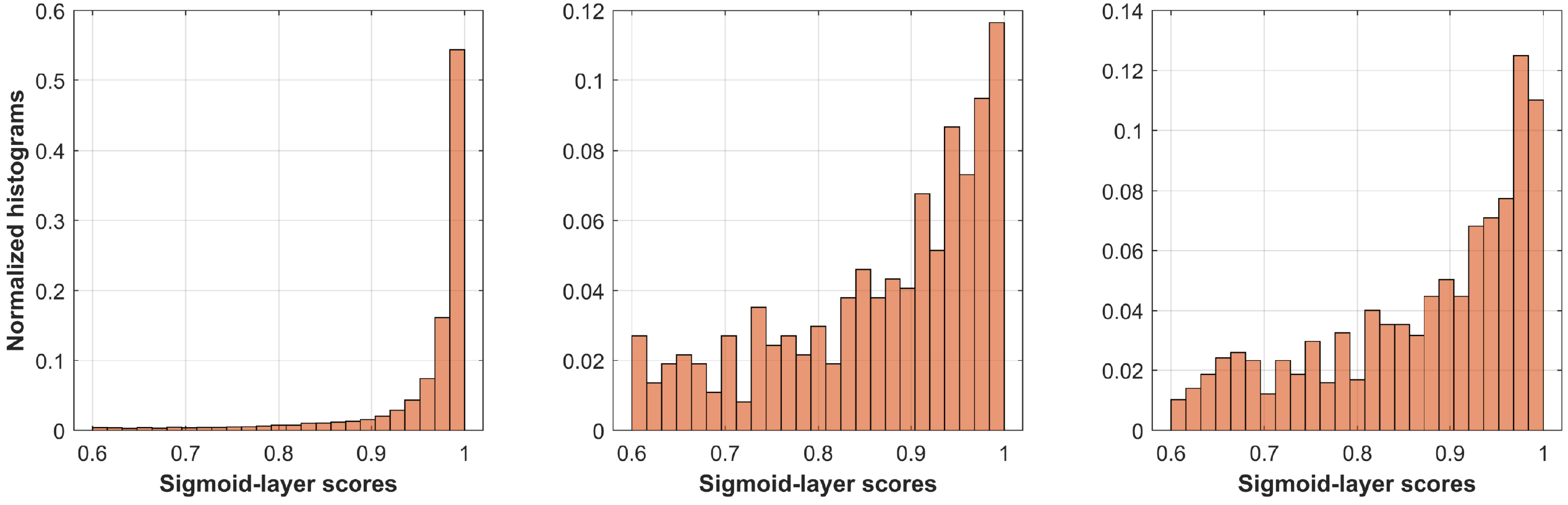}
			\subcaption{Score distributions of the true positive objects.}
			\label{TP_YOLOV4}
		\end{minipage}
		\hfill
		\vspace{0.35cm}
		\begin{minipage}[!t]{0.48\textwidth}
			\includegraphics[width=\textwidth]{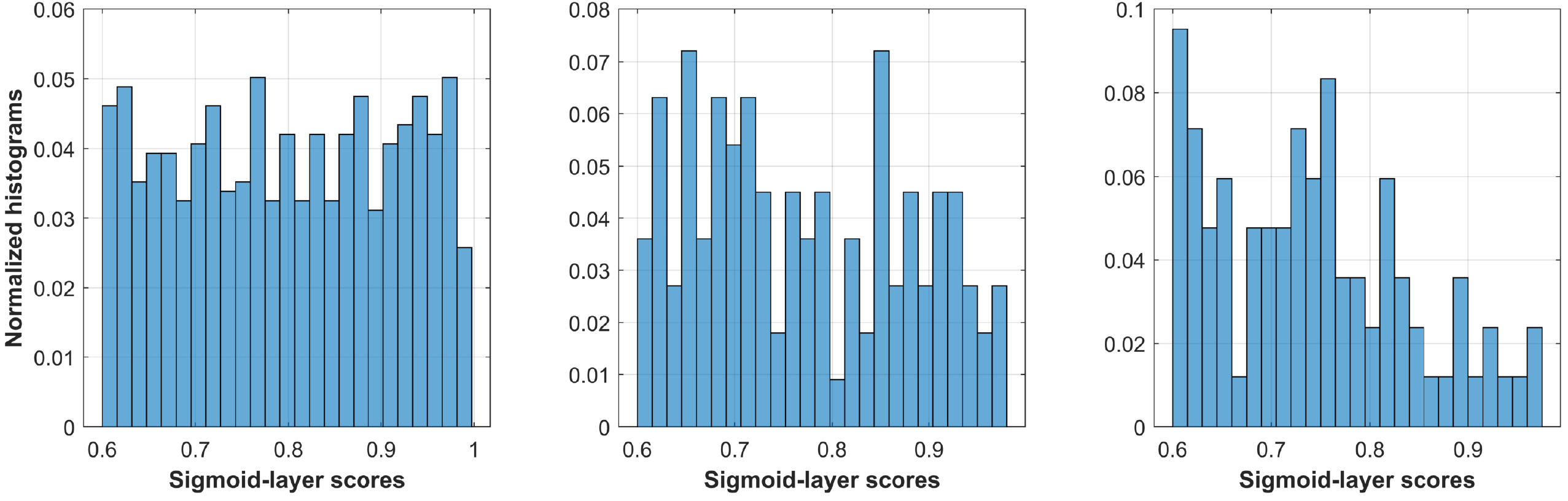}
			\subcaption{Score distributions of the false positive objects.}
			\label{FP_YOLOV4}
		\end{minipage}
		\caption{Distributions of the YOLOV4's classification scores for car, cyclist, and pedestrian classes, considering RGB modality.}
		\label{distribution_YOLOV4}
	\end{center}
\end{figure}

Even though YOLOV4 considers strategies to reduce overconfident predictions, our results demonstrate that a significant number of false positives are predicted with high score values, which demonstrates that the prediction layer using the Sigmoid function did not mitigate overconfident results enough, as shown in Fig. \ref{distribution_YOLOV4}.
\noindent
\raggedbottom

For object detection with $3D$ point clouds, we choose the lightweight yet effective SECOND \cite{yan2018SECOND} detector as the baseline. SECOND extracts features by encoding voxel-based $3D$ data with submanifold sparse $3D$ convolution layers \cite{yan2018SECOND}. The $3D$ features are converted to Bird's Eye View (BEV) re\-pre\-sentations via high compression, where the height in the metric space is flattened into the feature channels. Standard $2D$ convolutions are used to generate BEV features. The outputting feature map is passed to the single-stage anchor-based detector head for classification and bounding box regression. Compared to the sophisticated models with more structure information, the voxel-based SECOND \cite{yan2018SECOND} has a much faster runtime with comparable performance.

\begin{figure}[t]
	\begin{center}
		\begin{minipage}[t]{0.48\textwidth}
			\includegraphics[width=\textwidth]{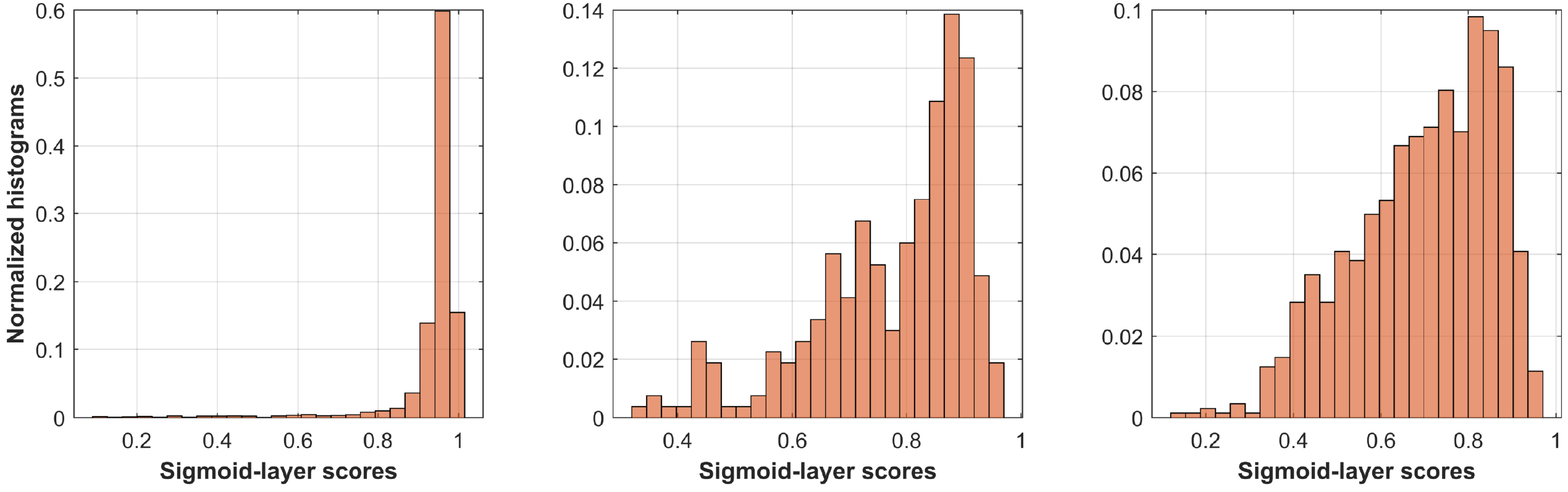}
			\subcaption{Score distributions of the true positive objects.}
			\label{TP_SECOND}
		\end{minipage}
		\hfill
		\vspace{0.35cm}
		\begin{minipage}[t]{0.48\textwidth}
			\includegraphics[width=\textwidth]{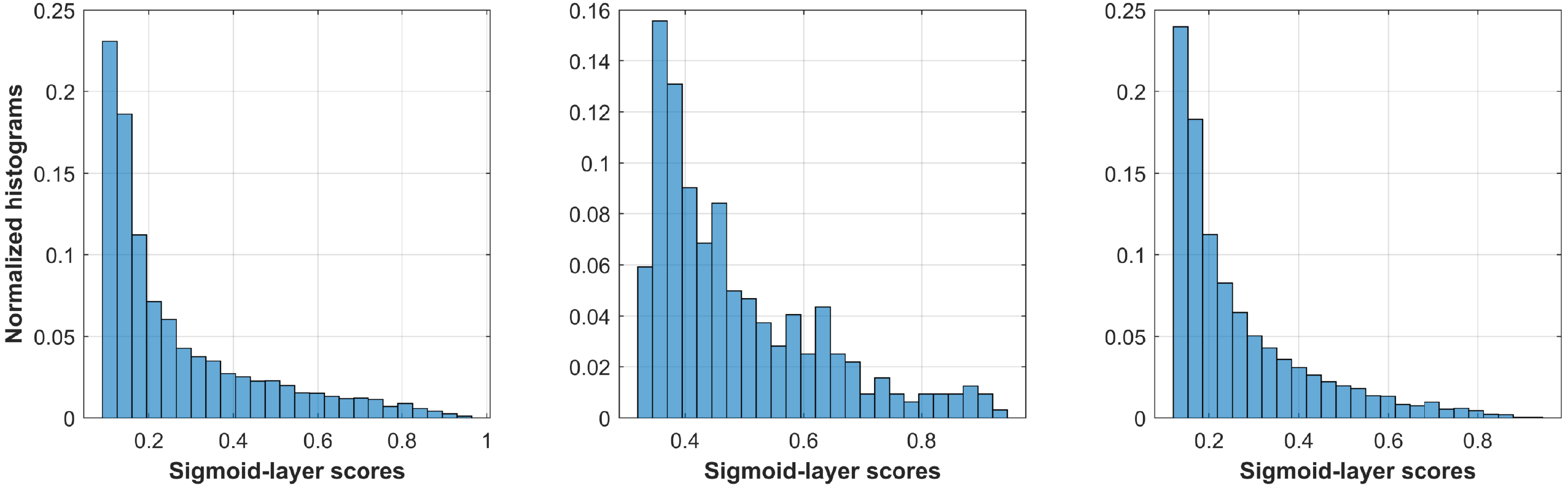}
			\subcaption{Score distributions of the false positive objects.}
			\label{FP_SECOND}
		\end{minipage}
		\caption{Distributions of the SECOND's classification scores for car, cyclist, and pedestrian classes, considering LiDAR modality ($3D$ LiDAR).}
		\label{distribution_SECOND}
	\end{center}
\end{figure}

As shown in Fig. \ref{distribution_SECOND}, SECOND \cite{yan2018SECOND} outputs a similar distribution, in a \textit{lato sensu} perspective, of the true positives as YOLOV4, while giving distinct and more ``aggressive" decisions on the false positives.

\noindent
\raggedbottom

\subsection{RGB and LiDAR Modalities}

\begin{figure}[!t]
	\begin{center}
		\begin{minipage}[t]{0.48\textwidth}
			\includegraphics[width=\textwidth]{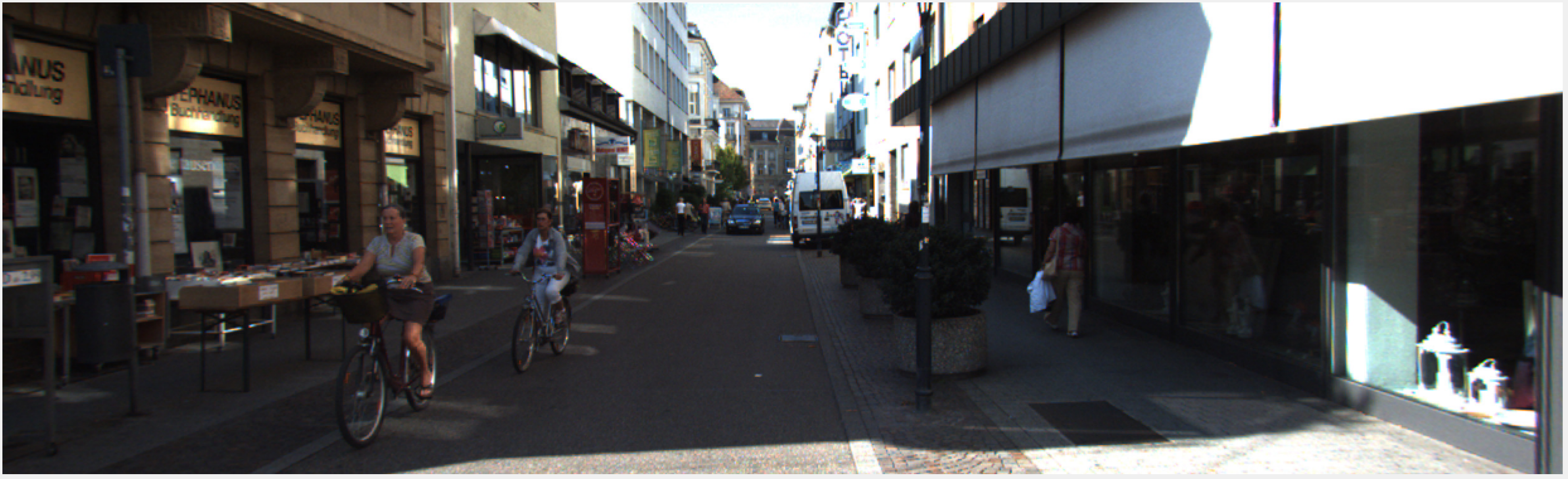}
			\subcaption{RGB modality.}
			\label{Frame007469}
		\end{minipage}
		\hfill
		\vspace{0.35cm}
		\begin{minipage}[!t]{0.48\textwidth}
			\includegraphics[width=\textwidth]{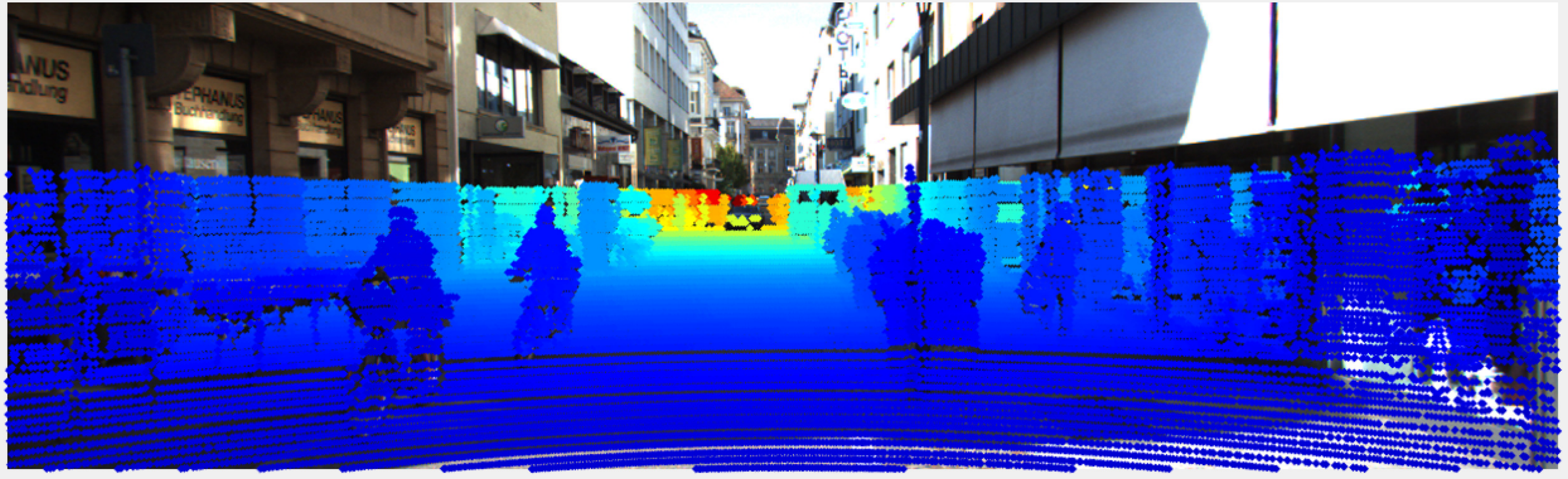}
			\subcaption{Projection of the $3D$ point clouds in the $2D$ image plain.}
			\label{Frame007469Projection}
		\end{minipage}
		\caption{The $3D$ cloud points were obtained from the Velodyne 64 sensor and then projected onto the image plane.}
		\label{RGB_Projection}
	\end{center}
\end{figure}\noindent
\raggedbottom
\begin{figure}[!t]
	\begin{center}
		\begin{minipage}[!t]{0.48\textwidth}
			\includegraphics[width=\textwidth]{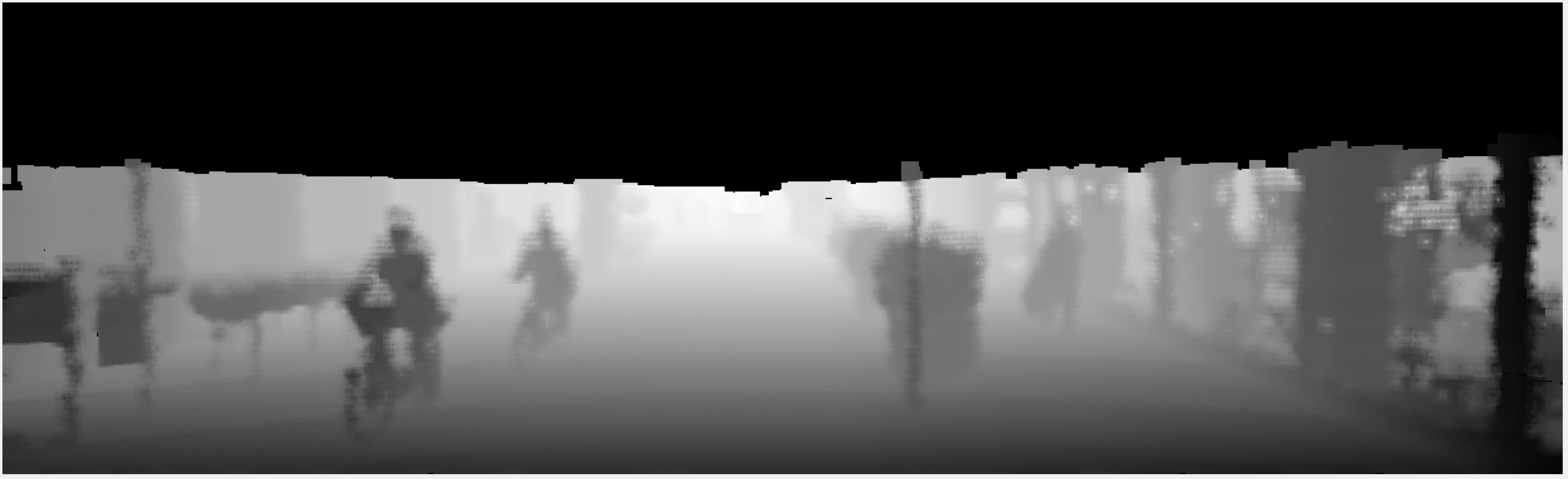}
			\subcaption{RaV map generated from LiDAR's depth data.}
			\label{RaV}
		\end{minipage}
		\hfill
		\vspace{0.35cm}
		\begin{minipage}[!t]{0.48\textwidth}
			\includegraphics[width=\textwidth]{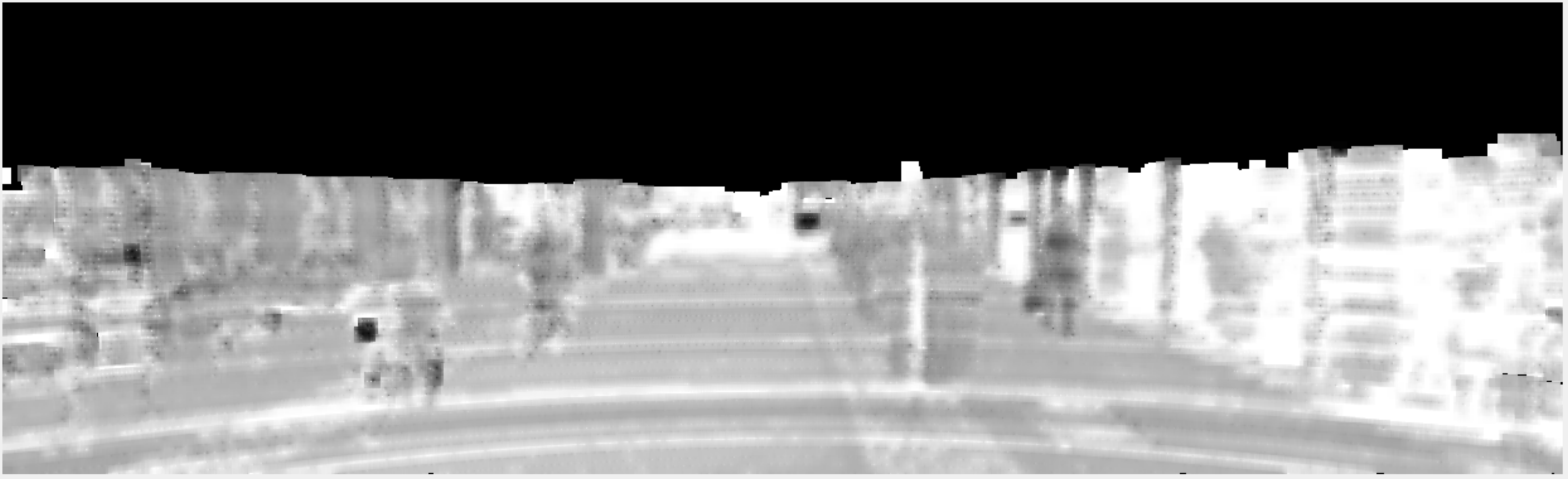}
			\subcaption{ReV map using the LiDAR's reflectance data.}
			\label{ReV}
		\end{minipage}
		\caption{Maps generated by bilateral filtering using sliding window with size $13\times 13$.}
		\label{RaV_ReV}
	\end{center}
\end{figure}\noindent
\raggedbottom

The proposed probabilistic methodology is validated through multi-sensory $2D$ and $3D$ object detection on the KITTI dataset, considering for YOLOV4 detector RGB images, range-view (RaV), and reflectance-view (ReV) maps modalities, as showed in Fig. \ref{RGB_Projection}, and $3D$ point clouds for SECOND detector. The modalities (RaV), and (ReV) were obtained by projecting the $3D-LiDAR$ point clouds in the $2D$ image plane followed by an upsampling step using a tailored bilateral filter implementation, expressed in (\ref{bf}), where $\hat{r_0}$ is the upsampled pixel~\cite{melotti_icarsc}
\begin{align}
	\hat{r_0} &= \frac{1}{W}\sum_{i=1}^n G_{\sigma_s}(||c_0-c_i||)G_{\sigma_r}(|r_0-r_i|)r_i,
	\label{bf}
\end{align}
where ${\displaystyle W=\sum_{i=1}^n G_{\sigma_s}(||c_0-c_i||)G_{\sigma_r}(r_0-r_i)}$ is a scaling factor that ensures the output sums to one, $G_{\sigma_s}$ weights the point $c_i$ inversely proportional to a distance (we used the Euclidean distance), and $G_{\sigma_r}$ weights the sampled points from their intensity values $r_i$. $G_{\sigma_s}$ and $G_{\sigma_r}$ were considered to be of the form
\begin{align}
	G_{\sigma_s} &= \frac{1}{1+(||c_0-c_i||)}\label{bf_gs},\\
	G_{\sigma_r} &= \frac{1}{1+(|r_0-r_i|)}\label{bf_gr}.
\end{align}

In fact, the upsample is for estimating points at positions where there are no projected points. The estimate of such points can be performed by considering a mask $C_{mask}$ of size $c\times c$ pixels, and using the sliding window principle. The sampled point $\hat{r_0}$, located at the center of $C_{mask}$, is weighted by the number of neighboring points defined by the mask size \ie, the formulation combines the intensity and distance values of a pixels group which are inside the mask $C_{mask}$, being $c_0=(c_h,c_v)$ the mask center, which is the localization of interest, and $\hat{r_0}$ the value to be estimated at $c_0$ from the $r_i$ (RaV or ReV), where $c_h$ and $c_v$ are the positions in the horizontal and vertical directions respectively, as in Fig. \ref{RaV_ReV}.

\section{Experiments and Results}
\label{sec:experiments}

In this section, we evaluate quantitatively the proposed approach to reduce overconfident predictions through the \textit{ML} and \textit{MAP} layers, considering Gaussian distributions, and normalized histograms, to model the prior and likelihood respectively. The approach depends of some ``hyperparameters" that interfere in the results achieved by the ML and MAP layers. The additive smoothing $\lambda$ (c.f. Sect. \ref{subsec.ML_MAP}), the chosen densities \eg, the numbers of bins of the normalized histograms (described in Sect. \ref{subsec.pdf_prior} above), are design dependent parameters and hence are subjected to the problem in hands. Here, the choice of these parameters has been made experimentally. 

The experiments conducted in this Section to assess the proposed technique and to support comparison studies make use of the KITTI `Object Detection' dataset \footnote{http://www.cvlibs.net/datasets/kitti/eval\_3dobject.php}, both the RGB (camera) and the LiDAR modalities (necessary for the RaV, ReV, and $3D$ point cloud). We have split the original training set by considering $3367$ frames for training, $375$ for validation, and then the remaining $3739$ frames comprise the actual test set. RGB, RaV, and ReV modalities were trained with the same hyperparameters (learning rate, image size, anchors, strides, {IoU} threshold, etc.) for YOLOV4, while the $3D$ point clouds were trained directly via the SECOND detector.

\subsection{General Performance and Overconfidence}
\begin{figure}[!t]
	\centering
	\includegraphics[scale=0.33]{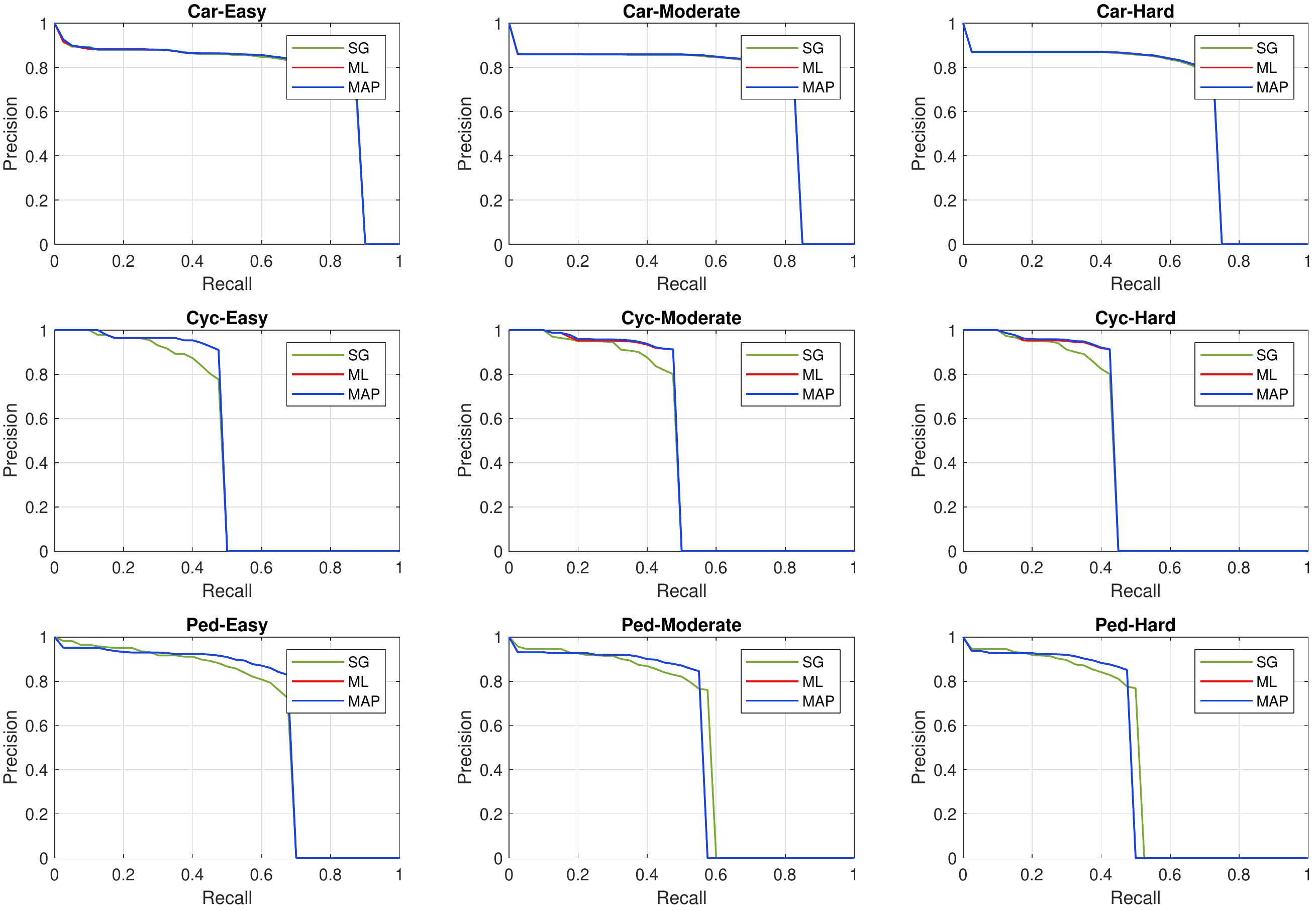}
	\caption{Precision-recall curves for car, cyc. and ped. classes using the RGB modality, with $\lambda_{ML}=1.6\times10^{-6}$, $Bins_{ML}=22$, $\lambda_{MAP}=1.0\times10^{-8}$, and $Bins_{MAP}=24$.}
	\label{Pr_Rc_rgb}
\end{figure}

\begin{figure}[!t]
	\centering
	\includegraphics[scale=0.33]{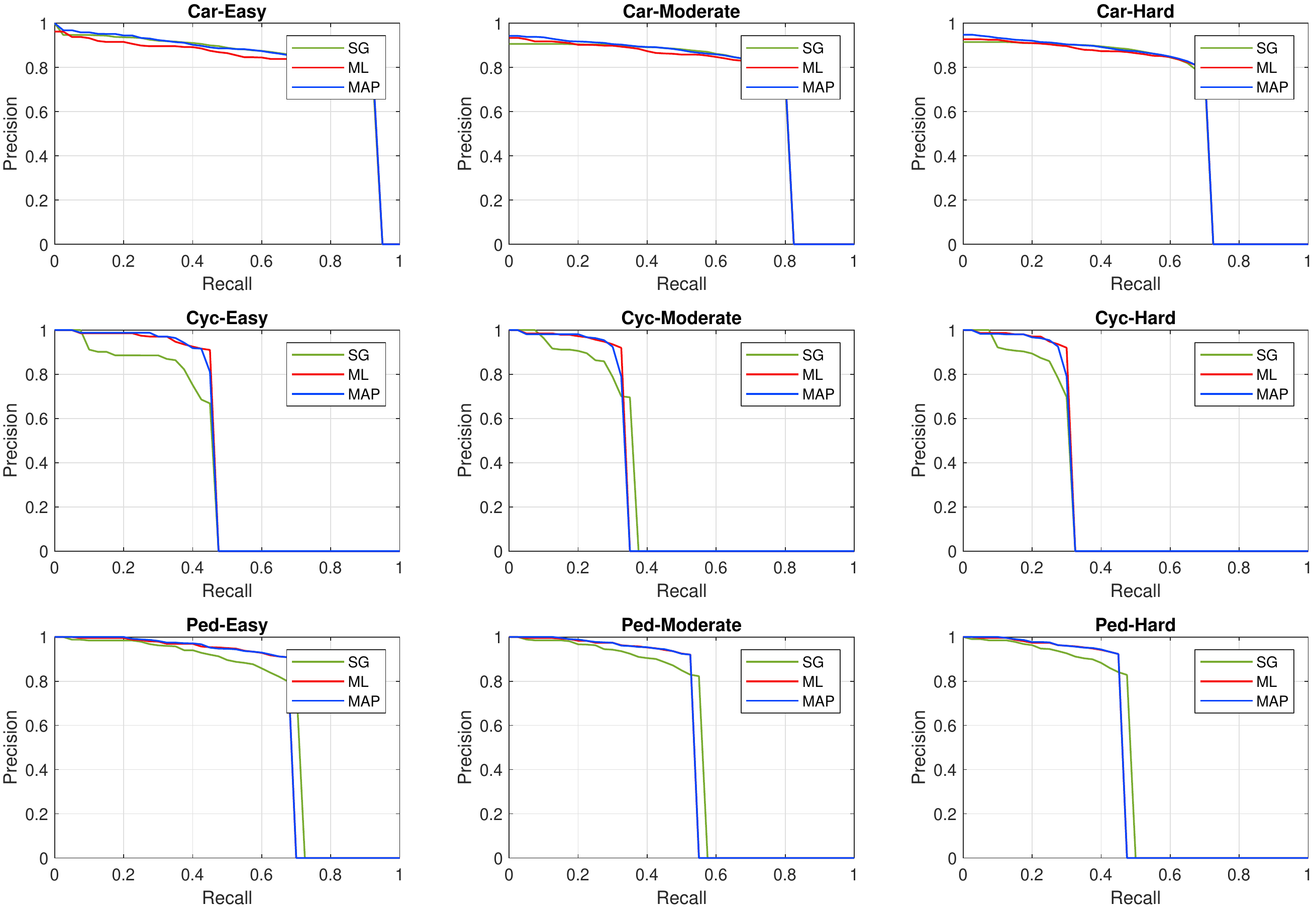}
	\caption{Precision-recall curves for RaV modality, with $\lambda_{ML}=1.3\times10^{-3}$, $Bins_{ML}=20$, $\lambda_{MAP}=1.7\times10^{-5}$, and $Bins_{MAP}=24$.}
	\label{Pr_Rc_dm}
\end{figure}

\begin{figure}[!t]
	\centering
	\includegraphics[scale=0.33]{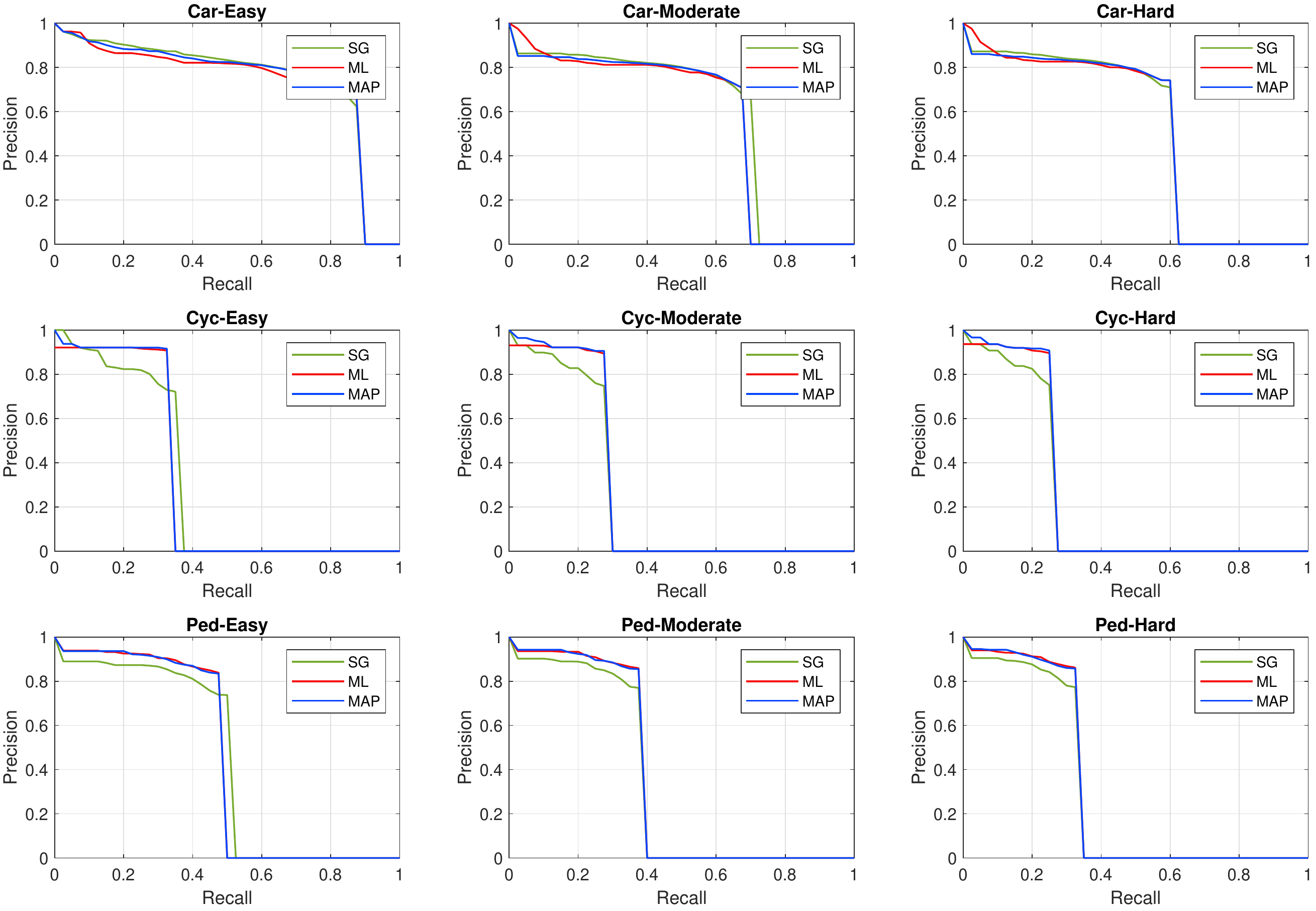}
	\caption{Precision-recall curves for ReV modality, with $\lambda_{ML}=1.3\times10^{-3}$, $Bins_{ML}=23$, $\lambda_{MAP}=8.0\times10^{-5}$, and $Bins_{MAP}=5$.}
	\label{Pr_Rc_rm}
\end{figure}

\begin{figure}[!t]
	\centering
	\includegraphics[scale=0.33]{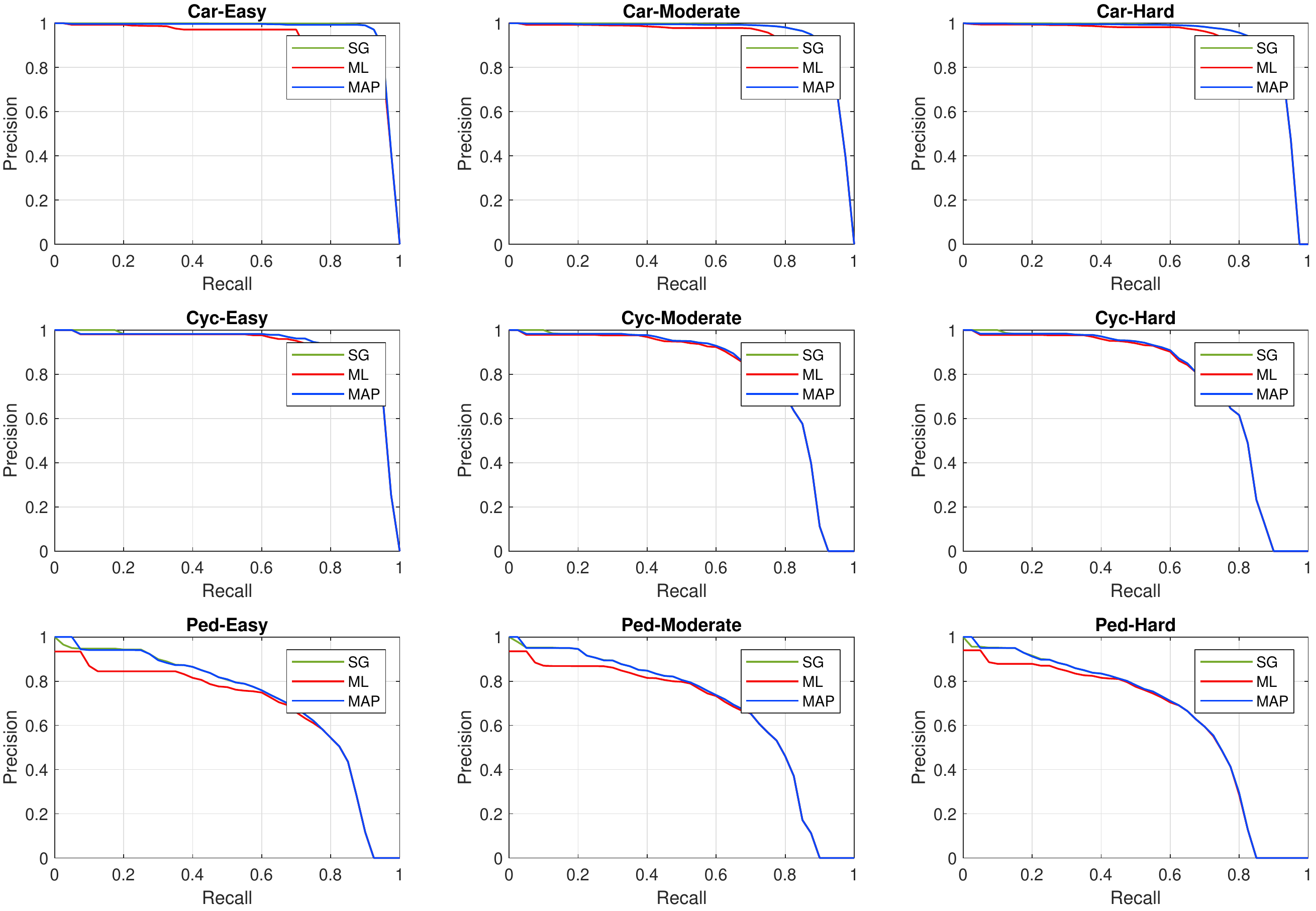}
	\caption{Precision-recall curves considering $2D$ bounding boxes after SECOND detector training.}
\label{Pr_Rc_Second2D}
\end{figure}

\begin{figure}[!t]
\centering
\includegraphics[scale=0.33]{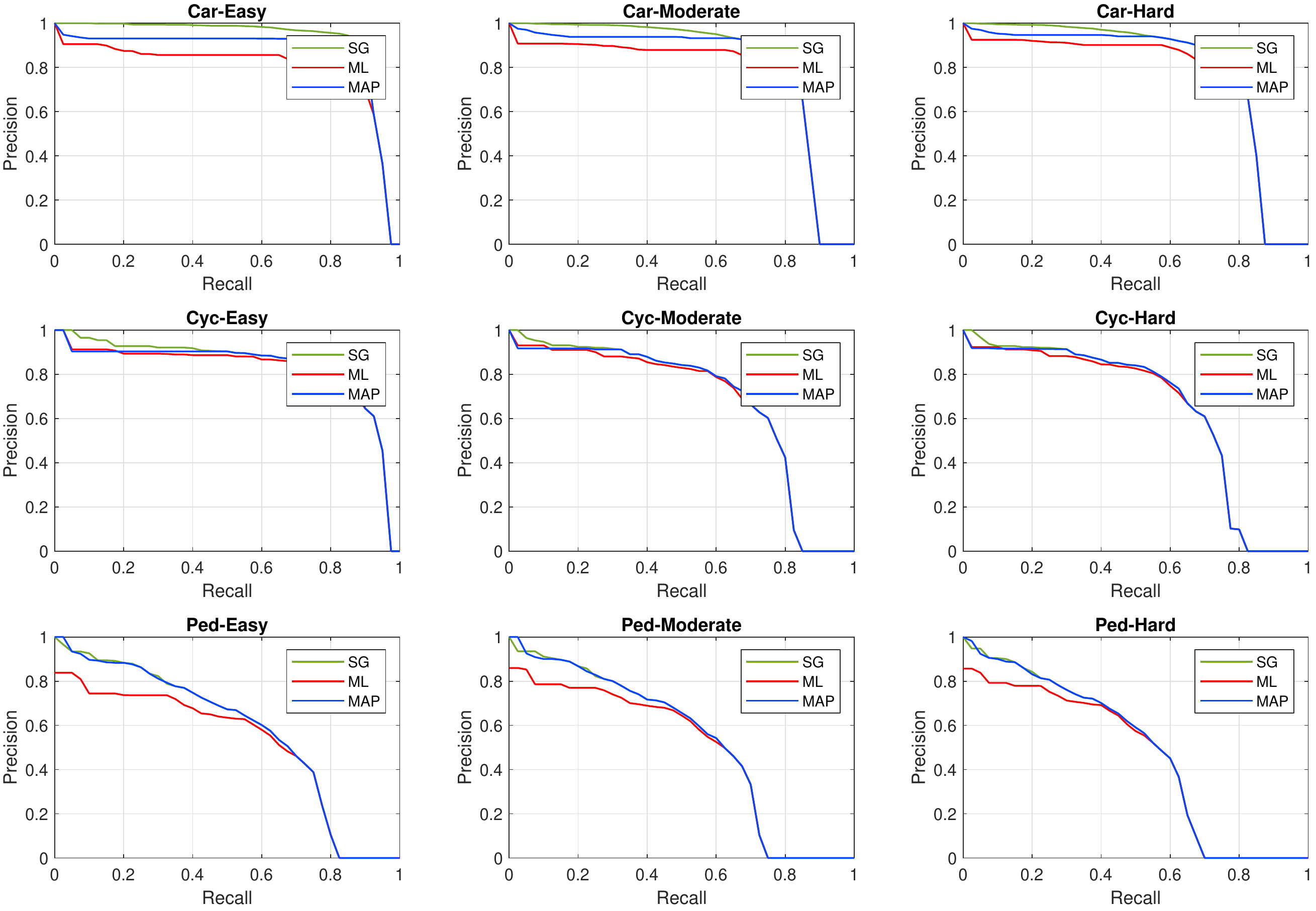}
\caption{Precision-recall curves using the SECOND detector to detect $3D$ bounding boxes.}
\label{Pr_Rc_Second3D}
\end{figure}

\begin{figure}[!t]
\centering
\includegraphics[scale=0.33]{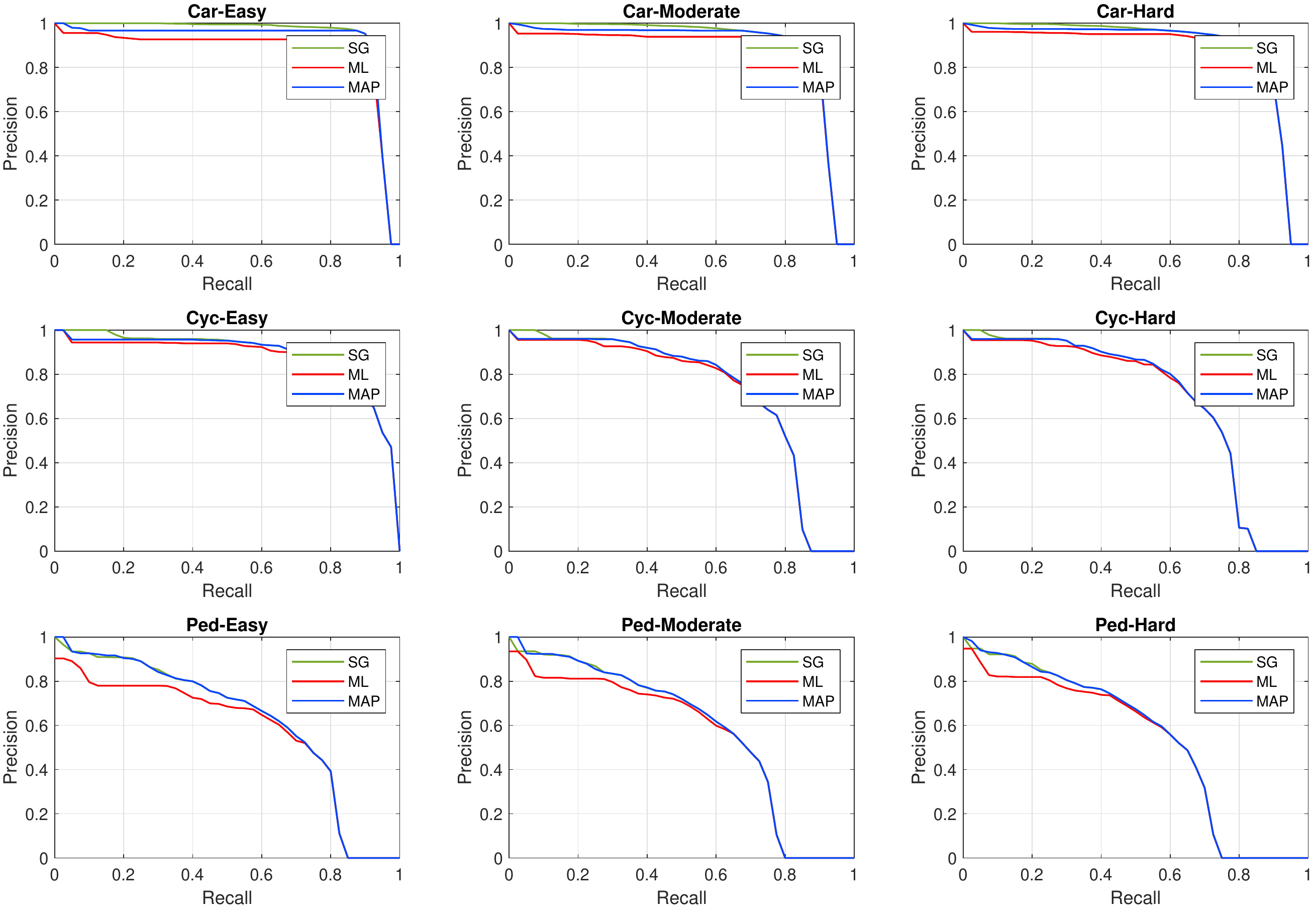}
\caption{Precision-recall curves considering $BEV$ detection.}
\label{Pr_Rc_SecondBEV}
\end{figure}

The results on the \textit{per-modalities} test sets are shown in figures \ref{Pr_Rc_rgb}, \ref{Pr_Rc_dm}, and \ref{Pr_Rc_rm} through precision-recall curves (Pr-Rc) for YOLOV4, while the figures \ref{Pr_Rc_Second2D}, \ref{Pr_Rc_Second3D}, and \ref{Pr_Rc_SecondBEV} correspond to the experimental results achieved with the SECOND detector. Note that the curves are presented to the three different difficulty levels (easy, moderate and hard), according to the KITTI dataset methodology for object detection.
 \noindent

 \noindent

 \noindent

 \noindent

 \noindent

 \noindent


In addition to the results given by the Pr-Rc curves, we further present a quantitative comparison, between the baseline (designated by Sigmoid, or simply \textit{SG}) and the proposed \textit{ML}, and \textit{MAP} layers, using the areas under the curve (AUC), as shown in tables \ref{auc_pre_rec_yolo} and \ref{auc_pre_rec_second}.
\begin{table*}[!t]
\begin{center}
\caption{AUC, in $\%$, for the baseline method denoted by \textit{SG}, and the proposed approaches (\textit{ML} and \textit{MAP} layers). The
results refer to the true positives and have been achieved by the YOLOV4 implementation using $2D$ representations.}
\begin{footnotesize}
\begin{tabular}{cccc|cccc|cccc}
	\toprule
	\multicolumn{12}{c}{\textbf{RGB Modality}}\\
	\multicolumn{4}{c}{\textbf{Easy}}          & \multicolumn{4}{c}{\textbf{Moderate}}        & \multicolumn{4}{c}{\textbf{Hard}}\\ 
	\hline \hline
	\textbf{Case} & \textbf{\textit{SG}}      & \textbf{\textit{ML}}      & \textbf{\textit{MAP}}     & \textbf{Case} & \textbf{\textit{SG}}      & \textbf{\textit{ML}}      & \textbf{\textit{MAP}}     & \textbf{Case} & \textbf{\textit{SG}}      & \textbf{\textit{ML}}      & \textbf{\textit{MAP}}     \\
	\textbf{Car}   & $75.48$ & $75.93$ & $\textbf{75.95}$ & \textbf{Car}   & $70.67$ & $70.90$ & $\textbf{71.00}$ & \textbf{Car}   & $63.04$ & $\textbf{63.36}$ & $\textbf{63.36}$ \\
	\textbf{Cyc}   & $45.47$ & $\textbf{47.20}$ & $\textbf{47.20}$ & \textbf{Cyc}   & $45.47$ & $46.83$ & $\textbf{46.99}$ & \textbf{Cyc}   & $40.94$ & $42.09$ & $\textbf{42.22}$ \\
	\textbf{Ped}   & $61.84$ & $\textbf{63.05}$ & $\textbf{63.05}$ & \textbf{Ped}   & $\textbf{52.27}$ & $51.25$ & $51.24$ & \textbf{Ped}   & $\textbf{45.65}$ & $44.52$ & $44.52$ \\
	\hline \hline
	\multicolumn{12}{c}{\textbf{RaV Modality}}\\
	\multicolumn{4}{c}{\textbf{Easy}}          & \multicolumn{4}{c}{\textbf{Moderate}}        & \multicolumn{4}{c}{\textbf{Hard}}\\ 
	\hline \hline
	\textbf{Case} & \textbf{\textit{SG}}      & \textbf{\textit{ML}}      & \textbf{\textit{MAP}}     & \textbf{Case} & \textbf{\textit{SG}}      & \textbf{\textit{ML}}      & \textbf{\textit{MAP}}     & \textbf{Case} & \textbf{\textit{SG}}      & \textbf{\textit{ML}}      & \textbf{\textit{MAP}}     \\
	\textbf{Car}   & $82.99$ & $81.13$ & $\textbf{83.21}$ & \textbf{Car}   & $71.07$ & $72.16$ & $\textbf{71.78}$ & \textbf{Car}   & $62.97$ & $62.80$ & $\textbf{63.53}$ \\
	\textbf{Cyc}   & $40.48$ & $\textbf{44.80}$ & $44.73$ & \textbf{Cyc}   & $32.28$ & $\textbf{32.74}$ & $32.43$ & \textbf{Cyc}   & $28.13$ & $\textbf{30.39}$ & $29.99$ \\
	\textbf{Ped}   & $66.27$ & $66.45$ & $\textbf{66.60}$ & \textbf{Ped}   & $\textbf{52.56}$ & $52.22$ & $52.22$ & \textbf{Ped}   & $\textbf{45.57}$ & $44.93$ & $44.96$ \\
	\hline \hline
	\multicolumn{12}{c}{\textbf{ReV Modality}}\\
	\multicolumn{4}{c}{\textbf{Easy}}          & \multicolumn{4}{c}{\textbf{Moderate}}        & \multicolumn{4}{c}{\textbf{Hard}}\\ 
	\hline \hline
	\textbf{Case} & \textbf{\textit{SG}}      & \textbf{\textit{ML}}      & \textbf{\textit{MAP}}     & \textbf{Case} & \textbf{\textit{SG}}      & \textbf{\textit{ML}}      & \textbf{\textit{MAP}}     & \textbf{Case} & \textbf{\textit{SG}}      & \textbf{\textit{ML}}      & \textbf{\textit{MAP}}     \\
	\textbf{Car}   & $\textbf{74.42}$ & $72.68$ & $73.92$ & \textbf{Car}   & $\textbf{58.13}$ & $56.14$ & $56.35$ & \textbf{Car}   & $50.83$ & $50.69$ & $50.52$ \\
	\textbf{Cyc}   & $30.80$ & $31.00$ & $\textbf{31.25}$ & \textbf{Cyc}   & $24.65$ & $26.46$ & $\textbf{26.86}$ & \textbf{Cyc}   & $22.73$ & $24.21$ & $\textbf{24.53}$ \\
	\textbf{Ped}   & $43.51$ & $\textbf{44.35}$ & $44.26$ & \textbf{Ped}   & $33.62$ & $35.44$ & $\textbf{35.45}$ & \textbf{Ped}   & $29.32$ & $\textbf{30.88}$ & $30.87$ \\
	\bottomrule
\end{tabular}
\end{footnotesize}
\label{auc_pre_rec_yolo}
\end{center}
\end{table*} \noindent \raggedbottom

\begin{figure}[t]
\begin{center}
\begin{minipage}[t]{0.48\textwidth}
\includegraphics[width=\textwidth]{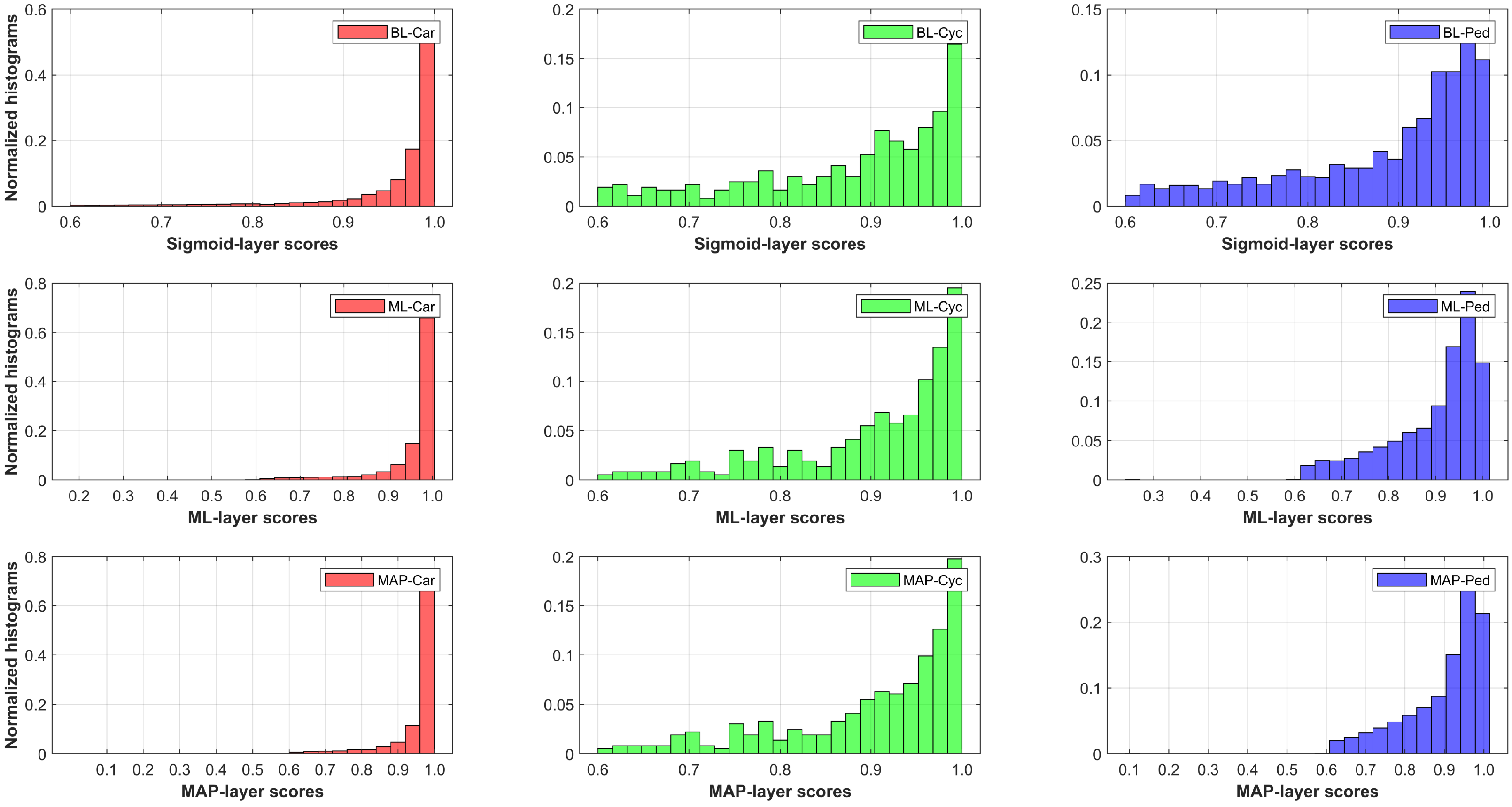}
\subcaption{RGB modality.}
\label{RGB_TP_Anexo_YOLO}
\end{minipage}
\\
\vspace{0.35cm}
\begin{minipage}[t]{0.48\textwidth}
\includegraphics[width=\textwidth]{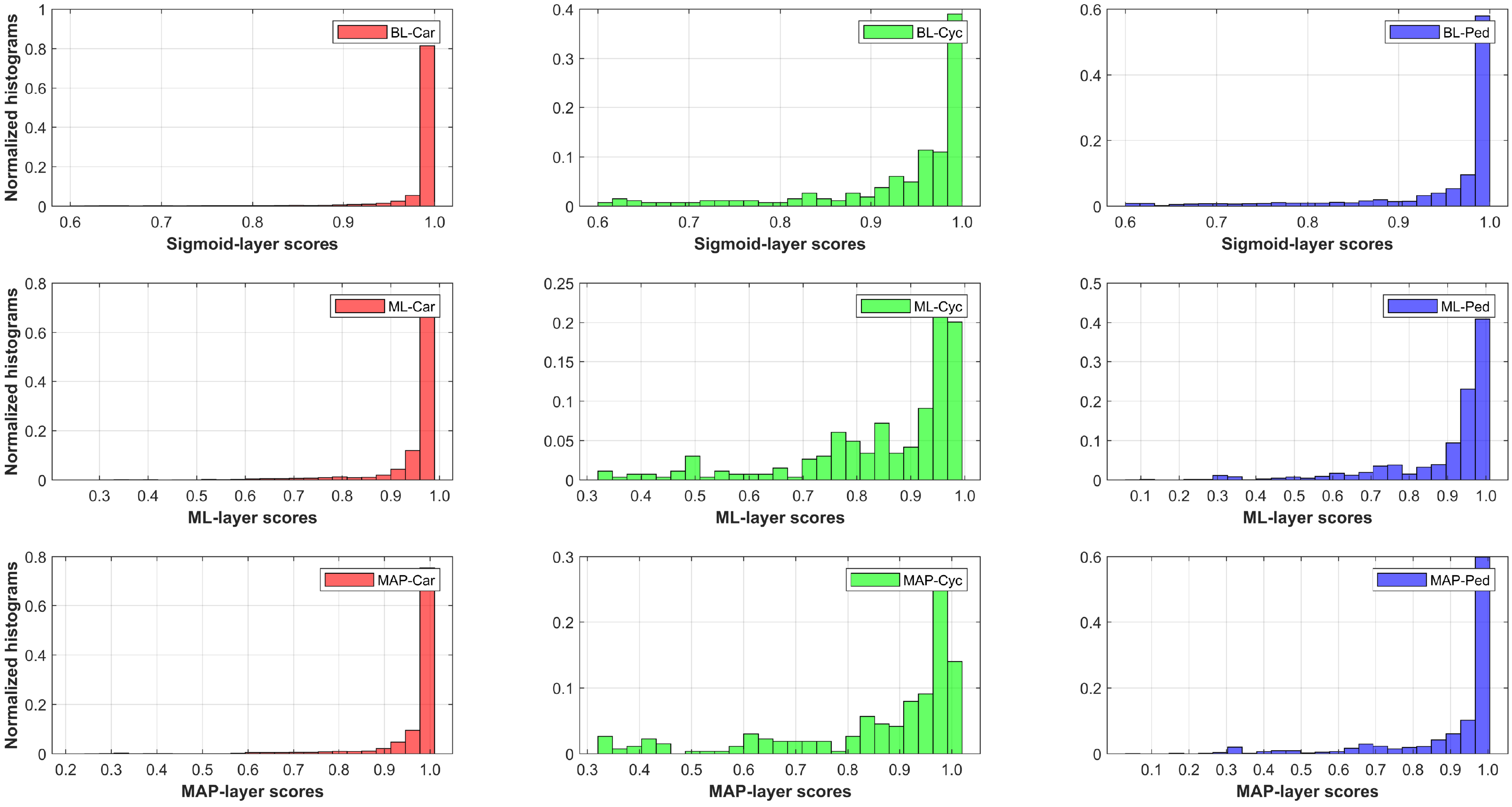}
\subcaption{RaV modality.}
\label{RaV_TP_Anexo_YOLO}
\end{minipage}
\\
\vspace{0.35cm}
\begin{minipage}[t]{0.48\textwidth}
\includegraphics[width=\textwidth]{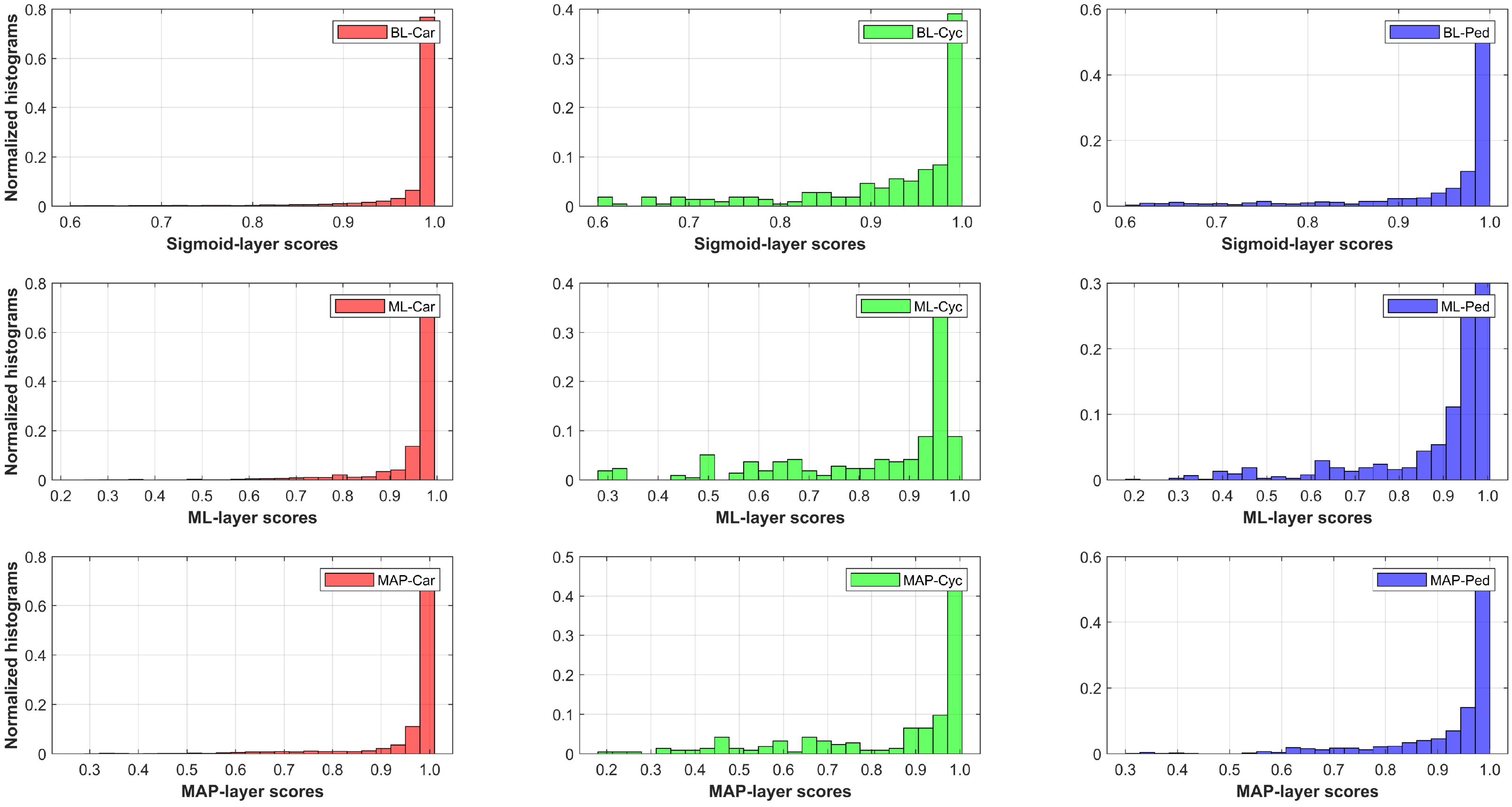}
\subcaption{ReV modality.}
\label{ReV_TP_Anexo_YOLO}
\end{minipage}
\caption{Score distributions considering TP objects from YOLOV4 detector.} \vspace*{3.in}
\label{RGB_RaV_ReV_TP_Anexo_YOLO}
\end{center}
\end{figure}

\begin{figure}[t]
\begin{center}
\begin{minipage}[t]{0.48\textwidth}
\includegraphics[width=\textwidth]{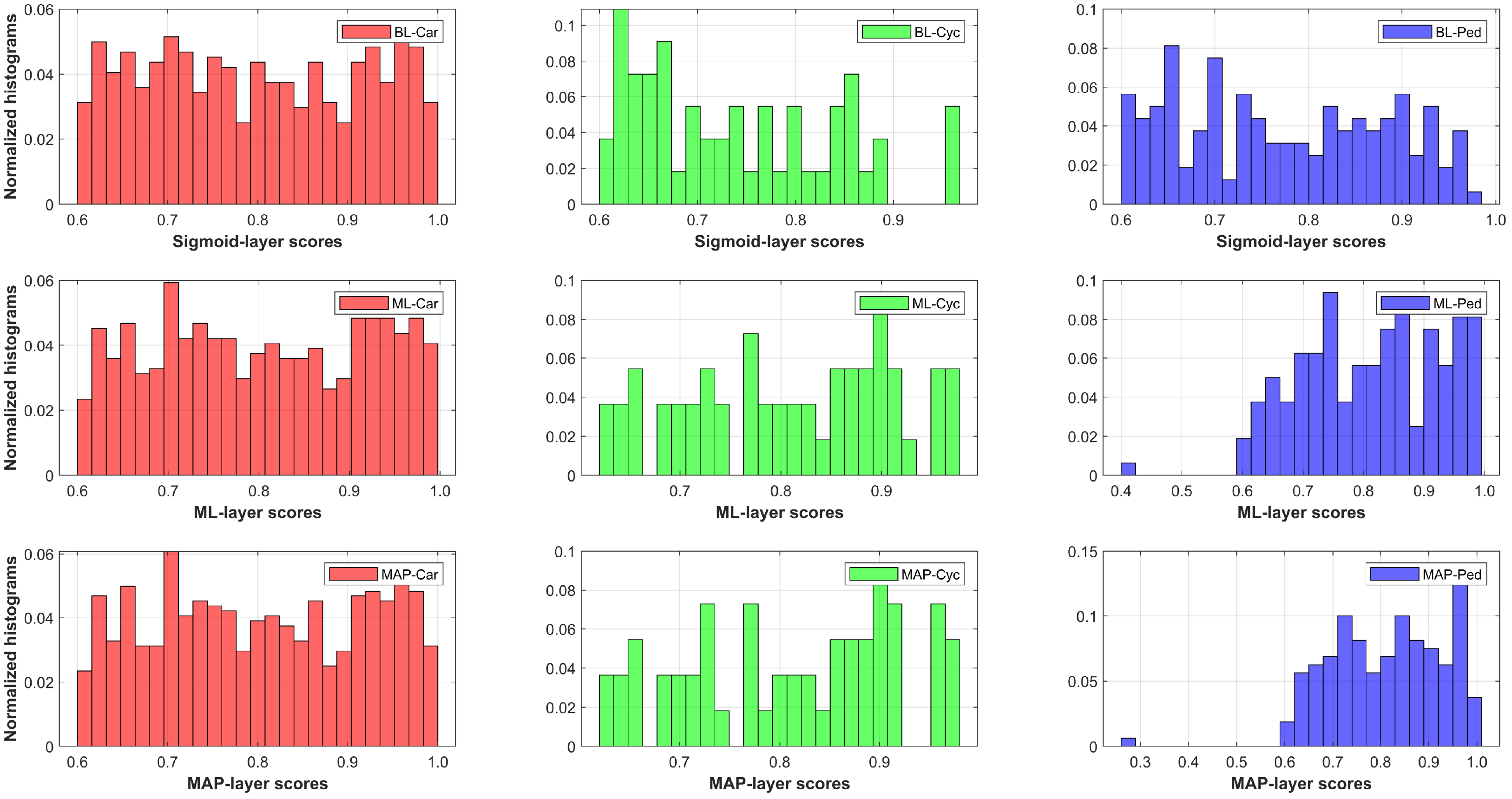}
\subcaption{RGB modality.}
\label{RGB_FP_Anexo_YOLO}
\end{minipage}
\\
\vspace{0.35cm}
\begin{minipage}[t]{0.48\textwidth}
\includegraphics[width=\textwidth]{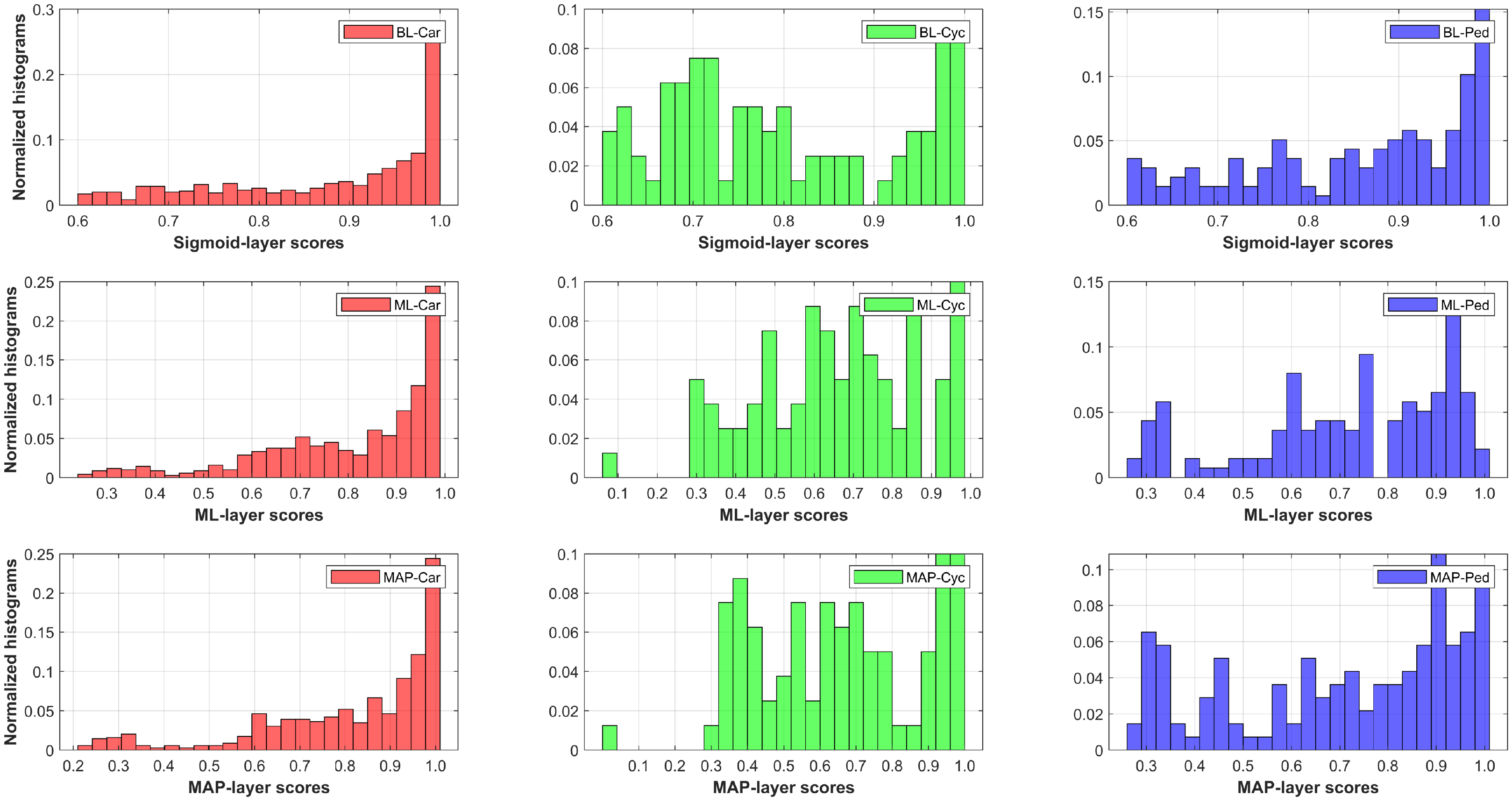}
\subcaption{RaV modality.}
\label{RaV_FP_Anexo_YOLO}
\end{minipage}
\\
\vspace{0.35cm}
\begin{minipage}[t]{0.48\textwidth}
\includegraphics[width=\textwidth]{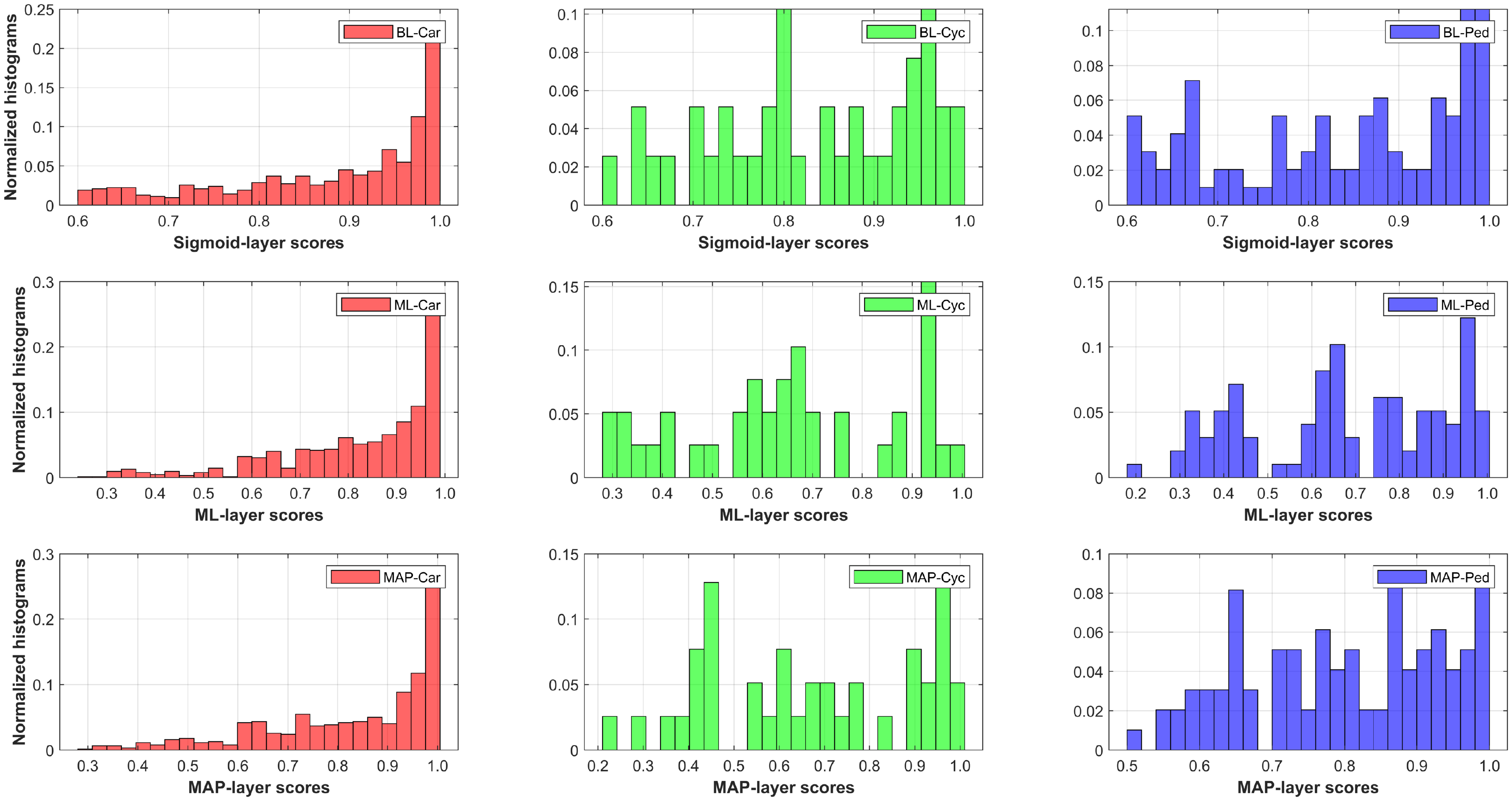}
\subcaption{ReV modality.}
\label{ReV_FP_Anexo_YOLO}
\end{minipage}
\caption{Score distributions considering FP objects from YOLOV4 detector.} \vspace*{3.in}
\label{RGB_RaV_ReV_FP_Anexo_YOLO}
\end{center}
\end{figure}

Based on the Pr-Rc curves using YOLOV4, it is possible to observe that the proposed probabilistic inference (\textit{ML}, and \textit{MAP} layers) outperformed the baseline (\textit{$SG$} layer) in almost all modalities and for most of the difficulty levels, particularly for the cyclist class, which has the smallest amount of objects in both training and test sets. To facilitate the comparison analysis, Table \ref{auc_pre_rec_yolo} contains the AUC from these experiments, where the best achieved detection performances are highlighted in bold. The AUC metrics show that ML and MAP achieved very satisfactory performance for different levels of difficulties and classes, as well as for different modalities. \CP{Furthermore}Additionally, the graphs in figures \ref{RGB_RaV_ReV_TP_Anexo_YOLO} and \ref{RGB_RaV_ReV_FP_Anexo_YOLO} show, when using the YOLOV4 detector, the distribution of the output-scores for the proposed approach and the baseline (\ie, using Sigmoid). We can see that \CP{we the values of the output-scores from the proposed approach are summarized by histograms shown in the Figures \ref{RGB_RaV_ReV_TP_Anexo_YOLO}, and \ref{RGB_RaV_ReV_FP_Anexo_YOLO} for the YOLOV4 detector} the baseline results achieved by YOLOV4 (shown in the first row) present many false positives (FP) with overconfident scores, while the \textit{ML} and \textit{MAP} layers have reduced the overconfidence on the FPs, whereas the performance on the true positives (TP) is relatively unaffected, according to Table \ref{TP_FP_yolov4}.
\noindent
\raggedbottom

\noindent
\raggedbottom

\begin{table*}[t]
\begin{center}
\caption{The average of the scores after the proposed approach, considering the results from the YOLOV4.}
\begin{tabular}{c|cccccccccc}
\toprule
\multirow{4}{*}{\begin{tabular}[c]{@{}c@{}}\textbf{True}\\ \textbf{Positives}\end{tabular}} & \textbf{Modality} & \multicolumn{3}{c}{\textbf{RGB}} & \multicolumn{3}{c}{\textbf{RaV}} & \multicolumn{3}{c}{\textbf{ReV}} \\ \cline{2-11}
&
\multicolumn{1}{c}{Approach} &
\multicolumn{1}{c}{\textit{SG}} &
\multicolumn{1}{c}{\textit{ML}} &
\multicolumn{1}{c|}{\textit{MAP}} &
\multicolumn{1}{c}{\textit{SG}} &
\multicolumn{1}{c}{\textit{ML}} &
\multicolumn{1}{c|}{\textit{MAP}} &
\multicolumn{1}{c}{\textit{SG}} &
\multicolumn{1}{c}{\textit{ML}} &
\multicolumn{1}{c}{\textit{MAP}} \\ \cline{2-11} 
&
\multicolumn{1}{c}{Average} &
\multicolumn{1}{c}{$0.947$} &
\multicolumn{1}{c}{$0.950$} &
\multicolumn{1}{c|}{$0.950$} &
\multicolumn{1}{c}{$0.974$} &
\multicolumn{1}{c}{$0.940$} &
\multicolumn{1}{c|}{$0.955$} &
\multicolumn{1}{c}{$0.970$} &
\multicolumn{1}{c}{$0.934$} &
\multicolumn{1}{c}{$0.951$} \\ 
&
\multicolumn{1}{c}{Variance} &
\multicolumn{1}{c}{$0.007$} &
\multicolumn{1}{c}{$0.006$} &
\multicolumn{1}{c|}{$0.006$} &
\multicolumn{1}{c}{$0.004$} &
\multicolumn{1}{c}{$0.010$} &
\multicolumn{1}{c|}{$0.011$} &
\multicolumn{1}{c}{$0.005$} &
\multicolumn{1}{c}{$0.011$} &
\multicolumn{1}{c}{$0.012$} \\ \hline\hline
\multirow{4}{*}{\begin{tabular}[c]{@{}c@{}}\textbf{False}\\ \textbf{Positives}\end{tabular}} & \textbf{Modality} & \multicolumn{3}{c}{\textbf{RGB}} & \multicolumn{3}{c}{\textbf{RaV}} & \multicolumn{3}{c}{\textbf{ReV}} \\ \cline{2-11}
&
\multicolumn{1}{c}{Approach} &
\multicolumn{1}{c}{\textit{SG}} &
\multicolumn{1}{c}{\textit{ML}} &
\multicolumn{1}{c|}{\textit{MAP}} &
\multicolumn{1}{c}{\textit{SG}} &
\multicolumn{1}{c}{\textit{ML}} &
\multicolumn{1}{c|}{\textit{MAP}} &
\multicolumn{1}{c}{\textit{SG}} &
\multicolumn{1}{c}{\textit{ML}} &
\multicolumn{1}{c}{\textit{MAP}} \\ \cline{2-11} 
&
\multicolumn{1}{c}{Average} &
\multicolumn{1}{c}{$0.788$} &
\multicolumn{1}{c}{$0.806$} &
\multicolumn{1}{c|}{$0.806$} &
\multicolumn{1}{c}{$0.867$} &
\multicolumn{1}{c}{$0.780$} &
\multicolumn{1}{c|}{$0.786$} &
\multicolumn{1}{c}{$0.872$} &
\multicolumn{1}{c}{$0.795$} &
\multicolumn{1}{c}{$0.817$} \\ 
&
\multicolumn{1}{c}{Variance} &
\multicolumn{1}{c}{$0.013$} &
\multicolumn{1}{c}{$0.013$} &
\multicolumn{1}{c|}{$0.013$} &
\multicolumn{1}{c}{$0.015$} &
\multicolumn{1}{c}{$0.037$} &
\multicolumn{1}{c|}{$0.044$} &
\multicolumn{1}{c}{$0.014$} &
\multicolumn{1}{c}{$0.034$} &
\multicolumn{1}{c}{$0.030$} \\ 
\bottomrule
\end{tabular}
\label{TP_FP_yolov4}
\end{center}
\end{table*}\noindent \raggedbottom

\begin{figure}[!t]
\begin{center}
\begin{minipage}[t]{0.48\textwidth}
\includegraphics[width=\textwidth]{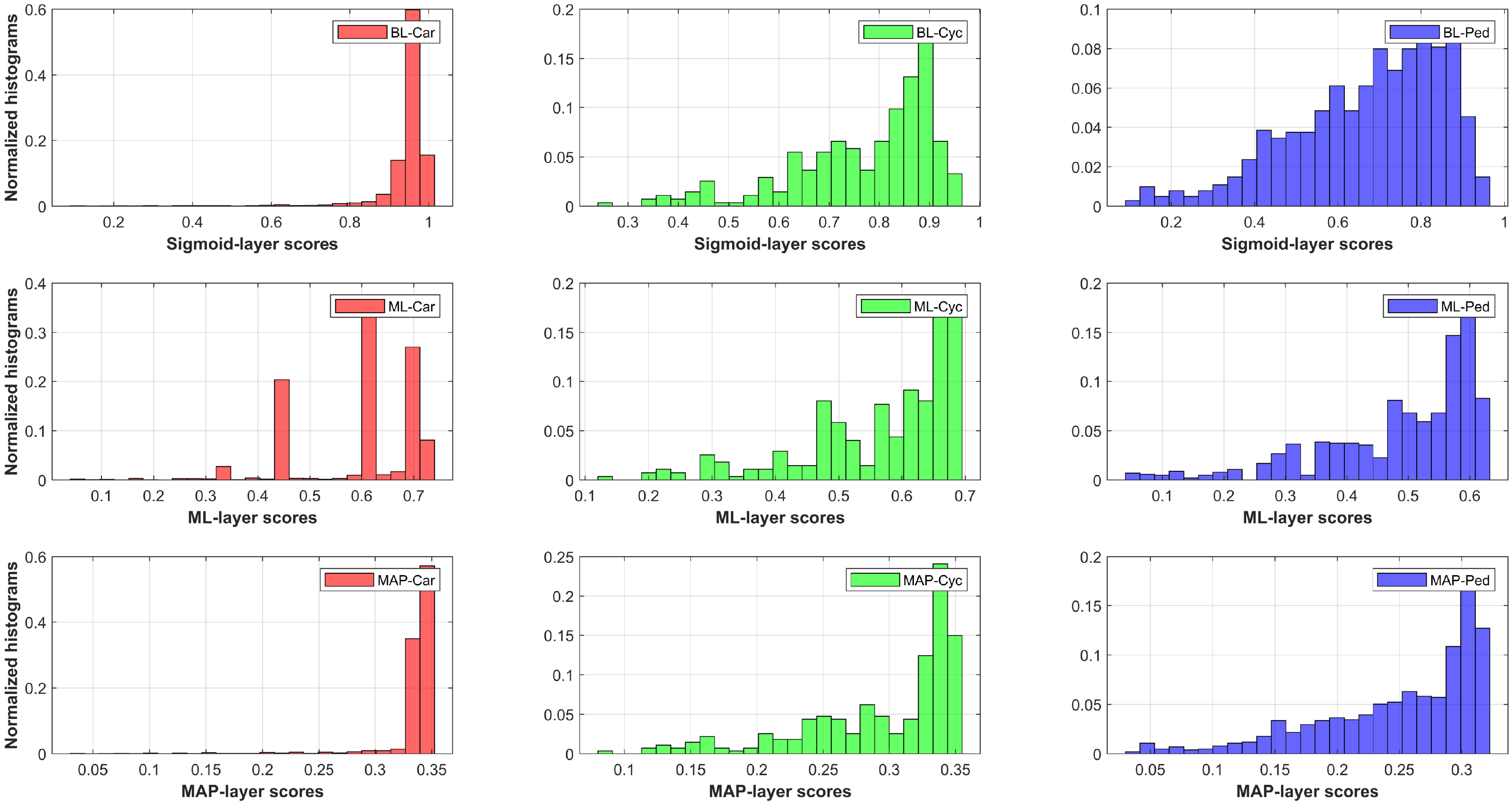}
\subcaption{TP objects.}
\label{Second2D_TP}
\end{minipage}
\\
\vspace{0.35cm}
\begin{minipage}[!t]{0.48\textwidth}
\includegraphics[width=\textwidth]{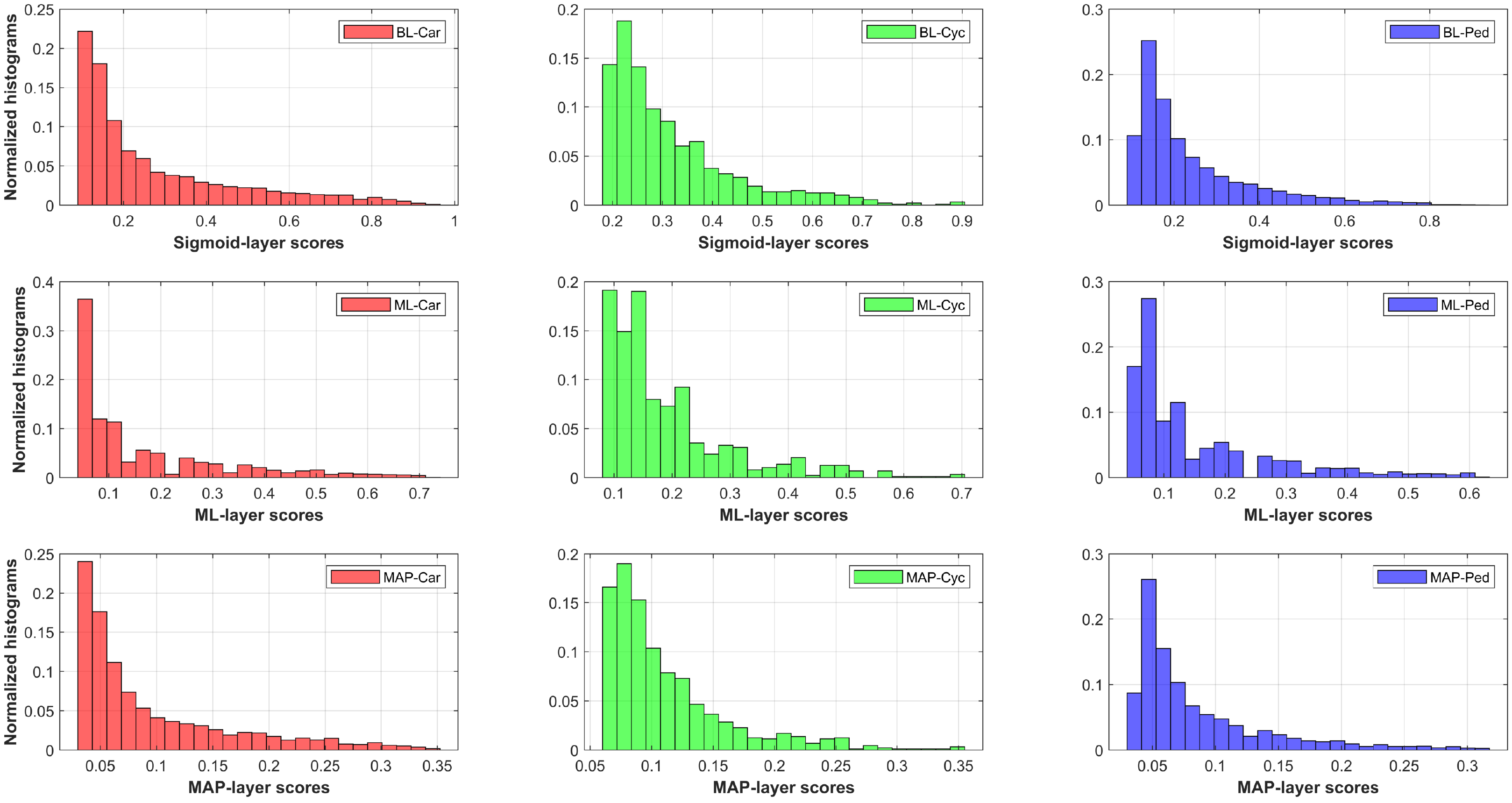}
\subcaption{FP objects.}
\label{Second3D_TP}
\end{minipage}
\caption{Score distributions considering objects from SECOND detector.} \vspace*{3.in}
\label{PC_TP_FP_Anexo_SECOND}
\end{center}
\end{figure}

The SECOND detector receives $3D$ point-clouds as input thus, besides $3D$ detection, we have converted the $3D$ representation to $2D$ and Bird's Eye View ($BEV$) for completeness of the results and benchmarking analysis. At first glance, the \textit{ML}, and \textit{MAP} approaches when applied to SECOND demonstrate to be less effective in improving the detection performance. This is due to the amount of high-scoring (\ie, highly confident) false positives is small in SECOND, as can be analyzed in Fig. \ref{PC_TP_FP_Anexo_SECOND} - this is more evident on the car category. Conversely, a bigger overlap of a relatively less distinguishable score range (0.4-0.6) can be improved by reweighing the scores. In this way, the probabilistic approach proposed in this work was applied to perform a `smoothing' on the classification scores to mitigate overconfidence, as can be seen from Fig. \ref{PC_TP_FP_Anexo_SECOND}, regarding the pedestrian class. Overall, we can say that the results achieved by the \textit{ML} and \textit{MAP} layers for the car and cyclist categories showed quite similar results compared to the baseline. Such results can be seen in Table \ref{auc_pre_rec_second}, this implies that the approach may compromise slightly the overall performance. The ML and MAP layers were compiled considering $\lambda_{ML}=5\times10^{-3}$, $Bins_{ML}=22$, $\lambda_{MAP}=1\times10^{-4}$, and $Bins_{MAP}=24$.
\begin{table*}[!t]
\begin{center}
\caption{AUC for the \textit{SG}, \textit{ML} and \textit{MAP} layers, using the SECOND detector, considering the true-positive objects.}
\begin{footnotesize}
\begin{tabular}{cccc|cccc|cccc}
	\toprule
	\multicolumn{12}{c}{\textbf{2D Detection}}\\
	\multicolumn{4}{c}{\textbf{Easy}}          & \multicolumn{4}{c}{\textbf{Moderate}}        & \multicolumn{4}{c}{\textbf{Hard}}\\ 
	\hline \hline
	\textbf{Case} & \textbf{\textit{SG}}      & \textbf{\textit{ML}}      & \textbf{\textit{MAP}}     & \textbf{Case} & \textbf{\textit{SG}}      & \textbf{\textit{ML}}      & \textbf{\textit{MAP}}     & \textbf{Case} & \textbf{\textit{SG}}      & \textbf{\textit{ML}}      & \textbf{\textit{MAP}}     \\
	\textbf{Car}   & $\textbf{96.88}$ & $93.09$ & $96.57$ & \textbf{Car}   & $\textbf{95.42}$ & $93.61$ & $95.24$ & \textbf{Car}   & $\textbf{93.02}$ & $91.89$ & $92.88$ \\
	\textbf{Cyc}   & $\textbf{92.66}$ & $91.91$ & $92.44$ & \textbf{Cyc}   & $\textbf{80.27}$ & $79.65$ & $80.14$ & \textbf{Cyc}   & $\textbf{76.65}$ & $76.11$ & $76.52$ \\
	\textbf{Ped}   & $70.77$ & $67.22$ & $\textbf{70.87}$ & \textbf{Ped}   & $67.74$ & $65.35$ & $\textbf{67.78}$ & \textbf{Ped}   & $64.09$ & $62.36$ & $\textbf{64.16}$ \\
	\hline \hline
	\multicolumn{12}{c}{\textbf{3D Detection}}\\
	\multicolumn{4}{c}{\textbf{Easy}}          & \multicolumn{4}{c}{\textbf{Moderate}}        & \multicolumn{4}{c}{\textbf{Hard}}\\ 
	\hline \hline
	\textbf{Case} & \textbf{\textit{SG}}      & \textbf{\textit{ML}}      & \textbf{\textit{MAP}}     & \textbf{Case} & \textbf{\textit{SG}}      & \textbf{\textit{ML}}      & \textbf{\textit{MAP}}     & \textbf{Case} & \textbf{\textit{SG}}      & \textbf{\textit{ML}}      & \textbf{\textit{MAP}}     \\
	\textbf{Car}   & $\textbf{91.80}$ & $79.40$ & $87.27$ & \textbf{Car}   & $\textbf{82.86}$ & $75.57$ & $80.32$ & \textbf{Car}   & $\textbf{79.86}$ & $75.16$ & $78.15$ \\
	\textbf{Cyc}   & $\textbf{84.21}$ & $81.89$ & $82.88$ & \textbf{Cyc}   & $\textbf{67.99}$ & $66.59$ & $67.31$ & \textbf{Cyc}   & $\textbf{64.03}$ & $62.80$ & $63.50$ \\
	\textbf{Ped}   & $\textbf{57.19}$ & $51.45$ & $57.11$ & \textbf{Ped}   & $52.39$ & $48.60$ & $\textbf{52.41}$ & \textbf{Ped}   & $\textbf{47.42}$ & $44.43$ & $47.38$ \\
	\hline \hline
	\multicolumn{12}{c}{\textbf{BEV Detection}}\\
	\multicolumn{4}{c}{\textbf{Easy}}          & \multicolumn{4}{c}{\textbf{Moderate}}        & \multicolumn{4}{c}{\textbf{Hard}}\\ 
	\hline \hline
	\textbf{Case} & \textbf{\textit{SG}}      & \textbf{\textit{ML}}      & \textbf{\textit{MAP}}     & \textbf{Case} & \textbf{\textit{SG}}      & \textbf{\textit{ML}}      & \textbf{\textit{MAP}}     & \textbf{Case} & \textbf{\textit{SG}}      & \textbf{\textit{ML}}      & \textbf{\textit{MAP}}     \\
	\textbf{Car}   & $\textbf{93.67}$ & $86.44$ & $91.55$ & \textbf{Car}   & $\textbf{89.81}$ & $85.64$ & $88.49$ & \textbf{Car}   & $\textbf{88.90}$ & $86.24$ & $88.02$ \\
	\textbf{Cyc}   & $\textbf{89.30}$ & $87.24$ & $88.59$ & \textbf{Cyc}   & $\textbf{72.41}$ & $71.17$ & $72.04$ & \textbf{Cyc}   & $\textbf{68.14}$ & $67.07$ & $67.85$ \\
	\textbf{Ped}   & $61.98$ & $57.26$ & $\textbf{62.07}$ & \textbf{Ped}   & $57.82$ & $54.83$ & $\textbf{57.89}$ & \textbf{Ped}   & $53.39$ & $51.11$ & $\textbf{53.41}$ \\
	\bottomrule
\end{tabular}
\end{footnotesize}
\label{auc_pre_rec_second}
\end{center}
\end{table*} \noindent \raggedbottom

The proposed technique for the SECOND detector tends to perform better on the `hard' level objects. We can conclude that, because the baseline implementation on SECOND does not attained overconfident behaviour, as shown by the results, the proposed approach degraded a bit the overall performance for that particular detector but, on the other hand, it smoothed the scores for the false positives (which is very desirable in autonomous driving), according to Table \ref{TP_FP_second}. Furthermore, the proposed approach has the advantage of giving probabilistic interpretation to the detectors.
\noindent
\raggedbottom

\begin{table*}[t]
\begin{center}
\caption{The average of the scores after the proposed approach, considering the results from the SECOND detector for $3D$ point clouds.}
\begin{tabular}{c|cccc|c|cccc}
\toprule
\multirow{3}{*}{\begin{tabular}[c]{@{}c@{}}True\\ Positives\end{tabular}} &
Approach & 
\textit{SG} &
\textit{ML} &
\textit{MAP} &
\multirow{3}{*}{\begin{tabular}[c]{@{}c@{}}False\\ Positives\end{tabular}} &
Approach &
\textit{SG} &
\textit{ML} &
\textit{MAP} \\ \cline{2-5} \cline{7-10}
&
Average &
$0.860$ &
$0.570$ &
$0.310$ &
&
Average &
$0.258$ &
$0.161$ &
$0.091$ \\
&
Variance &
$0.030$ &
$0.017$ &
$0.008$ &
&
Variance &
$0.026$ &
$0.017$ &
$0.005$
\\
\bottomrule
\end{tabular}
\label{TP_FP_second}
\end{center}
\end{table*}\noindent \raggedbottom

As the SECOND detector provides a relatively regularized scores across the classes, the \textit{ML} and \textit{MAP} approaches have limited improvement by eliminating the high-scoring FPs. However, the probabilistic approach is able to distinguish the ambiguous scores from the pedestrian class. This can be shown by the more overlap score range of true and false positive objects (Figures \ref{Pr_Rc_Second2D}, \ref{Pr_Rc_Second3D}, \ref{Pr_Rc_SecondBEV}, and \ref{PC_TP_FP_Anexo_SECOND}). 

\subsection{Calibration Error}

Typically, the calibration of probabilistic predictions (which relates the model's prediction scores to the true correctness likelihood~\cite{Niculescu}) is analyzed by the Expected Calibration Error (ECE) metric~\cite{oncalibration}. The ECE is obtained from a histogram with $M$ bins, where each bin contains a group of scores (predicted values). Each object with its respective classification score is allocated within a bin, according to the prediction confidence \ie, maximum prediction value. Each bin $B_m$ is defined through a range $I_m=\big (\frac{(m-1)}{M},\frac{m}{M}\big ]$, where $m={1,..,M}$. The average accuracy - $acc(B_m)$ - is obtained for each bin $B_m$, as well as the average confidence $conf(B_m)=\frac{1}{|B_m|}\sum_i\hat{p_i}$, where $\hat{p_i}$ is the confidence for classified object $i$ and $|B_m|$ is the amount of objects in each bin $B_m$. From the $acc(\cdot)$ and $conf(\cdot)$, the ECE is obtained according to (\ref{ece_metric}):
\begin{align}
ECE=\sum\limits_{m=1}^{M}\cfrac{|B_m|}{n}|acc(B_m)-conf(B_m)|,
\label{ece_metric}
\end{align}
where $n$ is the total the number of objects. Thus, the proposed approach can be compared quantitatively with the \CP{Sigmoid prediction function} baseline through the ECE, as shown in Table \ref{ece_yolov4} (RGB, RaV and ReV modalities) and Table \ref{ece_second} ($3D$ Point clouds). Based on the results shown in Table \ref{ece_yolov4}, considering the YOLOV4 detector, we can see that the ECE was reduced for the proposed methodology. However, for the SECOND detector applied to point-cloud representation, the achieved ECE remained close to the baseline - as shown in Table \ref{ece_second}.\\

\begin{table}[!t]
\begin{center}
\caption{ECE on the different modalities, when using YOLOV4 as detector.}
\begin{tabular}{cccc}
\toprule
\textbf{} & \multicolumn{3}{c}{\textbf{RGB Modality}} \\ 
\textbf{Method:} & \textbf{SG (baseline)} & \textbf{ML} & \textbf{MAP} \\
ECE & 0.007 & \textbf{0.005} & \textbf{0.005} \\
\hline
\textbf{} & \multicolumn{3}{c}{\textbf{RaV Modality}} \\
\textbf{Method:} & \textbf{SG (baseline)} & \textbf{ML} & \textbf{MAP} \\
ECE & 0.036 & \textbf{0.013} & 0.027 \\
\hline
\textbf{} & \multicolumn{3}{c}{\textbf{ReV Modality}} \\
\textbf{Method:} & \textbf{SG (baseline)} & \textbf{ML} & \textbf{MAP} \\
ECE & 0.031 & \textbf{0.013} & 0.031 \\
\bottomrule
\end{tabular}
\label{ece_yolov4}
\end{center}
\end{table}\noindent \raggedbottom

\begin{table}[!t]
\begin{center}
\caption{ECE for the detector SECOND - $3D$ point clouds.}
\begin{tabular}{cccc}
\toprule
\textbf{} & \multicolumn{3}{c}{\textbf{3D - PointCloud}} \\ 
\textbf{Method:} & \textbf{SG (baseline)} & \textbf{ML} & \textbf{MAP} \\
ECE & \textbf{0.196} & 0.323 & 0.208 \\
\bottomrule
\end{tabular}
\label{ece_second}
\end{center}
\end{table}\noindent \raggedbottom

\section{Concluding Remarks}

Many machine learning models, particularly deep learning ones, have the tendency of regarding the values of the detected objects' scores as being a degree of confidence (or related to a probability) without any level of uncertainty \ie, many deep models are not formulated to provide uncertainties associated with the predicted results. One way to ensure that the classification scores of detected objects can be interpreted as probabilistic values or have some level of uncertainty is through calibration/regularization techniques. However, the developments of such techniques are quite challenging, for instance because there is no ground truth available on uncertainty data for - and it is still an open problem.

The state-of-the-art formalism to capture model uncertainties (calibration/regularization techniques), during training or at the time test phase, aim to ensure confidence measures for the predictions of the models. In this way, this paper proposes a formulation considering the concepts of Maximum Likelihood (ML) and Maximum a-Posteriori (MAP) to reduce the overconfidence of detected false positive objects from the classification scores \ie, the ML/MAP layers are be able to reduce confidence in incorrect predictions. The formulation takes into account a probabilistic inference through two models, one being non-parametric (normalized histogram) and the other is parametric (Gaussian density to model the priors for the MAP).

As a way to present the efficiency of the proposed probabilistic inference approach, this work considered different modalities, as RGB imagens, RaV, and ReV maps, as well as $3D$ point clouds data \ie, datasets with different characteristics. In the case of RGB images, the characteristics are obtained directly from the camera, while RaV and ReV maps are obtained from depth (range-view) and intensity (reflectance-view) data, respectively. In addition, this paper has considered the detection of objects directly on 3D point clouds, as input, \CP{directly} processed by a LiDAR-based pipeline - SECOND \cite{yan2018SECOND}.

The results achieved by the proposed approach are very satisfactory, specially for the cyclists class (for YOLOV4), and pedestrian case (for SECOND), as evidenced by the improvements in general performance (evaluated with the Pr-Rc curves and AUC), reduction of overconfidence (illustrated in Figures \ref{RGB_RaV_ReV_TP_Anexo_YOLO}, \ref{RGB_RaV_ReV_FP_Anexo_YOLO} and
\ref{PC_TP_FP_Anexo_SECOND}) and a general reduction in the calibration error (evaluated using the ECE). Finally, a key advantage of the proposed approach is that there is no need to perform a new network training, that is, the approach has been applied in already trained networks.

\ifCLASSOPTIONcaptionsoff
\newpage
\fi

\clearpage 
\bibliographystyle{IEEEtran}
\bibliography{refs}

\begin{thebibliography}{10}
\providecommand{\url}[1]{#1}
\csname url@samestyle\endcsname
\providecommand{\newblock}{\relax}
\providecommand{\bibinfo}[2]{#2}
\providecommand{\BIBentrySTDinterwordspacing}{\spaceskip=0pt\relax}
\providecommand{\BIBentryALTinterwordstretchfactor}{4}
\providecommand{\BIBentryALTinterwordspacing}{\spaceskip=\fontdimen2\font plus
\BIBentryALTinterwordstretchfactor\fontdimen3\font minus
  \fontdimen4\font\relax}
\providecommand{\BIBforeignlanguage}[2]{{%
\expandafter\ifx\csname l@#1\endcsname\relax
\typeout{** WARNING: IEEEtran.bst: No hyphenation pattern has been}%
\typeout{** loaded for the language `#1'. Using the pattern for}%
\typeout{** the default language instead.}%
\else
\language=\csname l@#1\endcsname
\fi
#2}}
\providecommand{\BIBdecl}{\relax}
\BIBdecl

\bibitem{road}
G.~Singh, S.~Akrigg, M.~D. Maio, V.~Fontana, R.~J. Alitappeh, S.~Khan, S.~Saha,
  K.~Jeddisaravi, F.~Yousefi, J.~Culley, T.~Nicholson, J.~Omokeowa,
  S.~Grazioso, A.~Bradley, G.~D. Gironimo, and F.~Cuzzolin, ``Road: The road
  event awareness dataset for autonomous driving,'' \emph{IEEE Transactions on
  Pattern Analysis and Machine Intelligence}, vol.~45, no.~1, pp. 1036--1054,
  2023.

\bibitem{KITTI360}
Y.~Liao, J.~Xie, and A.~Geiger, ``Kitti-360: A novel dataset and benchmarks for
  urban scene understanding in 2d and 3d,'' \emph{IEEE Transactions on Pattern
  Analysis and Machine Intelligence}, vol.~45, no.~3, pp. 3292--3310, 2023.

\bibitem{HeQingdong}
Q.~He, Z.~Wang, H.~Zeng, Y.~Zeng, Y.~Liu, S.~Liu, and B.~Zeng, ``Stereo {RGB}
  and deeper lidar-based network for 3d object detection in autonomous
  driving,'' \emph{IEEE Transactions on Intelligent Transportation Systems},
  vol.~24, no.~1, pp. 152--162, 2023.

\bibitem{Janai2019}
J.~Janai, F.~Güney, A.~Behl, and A.~Geiger, ``Computer vision for autonomous
  vehicles: Problems, datasets and state of the art,'' \emph{Foundations and
  Trends in Computer Graphics and Vision}, vol.~12, no. 1–3, pp. 1--308,
  2020.

\bibitem{Shaoshan2017}
S.~Liu, L.~Li, J.~Tang, S.~Wu, and J.-L. Gaudiot, ``Creating autonomous vehicle
  systems,'' \emph{Synthesis Lectures on Computer Science}, vol.~6, no.~1, pp.
  i--186, 2017.

\bibitem{Claussmann}
L.~Claussmann, M.~Revilloud, D.~Gruyer, and S.~Glaser, ``A review of motion
  planning for highway autonomous driving,'' \emph{IEEE Transactions on
  Intelligent Transportation Systems}, vol.~21, no.~5, pp. 1826--1848, 2020.

\bibitem{RobotCarDataset}
W.~Maddern, G.~Pascoe, C.~Linegar, and P.~Newman, ``{1 Year, 1000km: The Oxford
  RobotCar Dataset},'' \emph{The International Journal of Robotics Research},
  vol.~36, no.~1, pp. 3--15, 2017.

\bibitem{as}
S.~Aly, ``Partially occluded pedestrian classification using histogram of
  oriented gradients and local weighted linear kernel support vector machine,''
  \emph{IET Computer Vision}, vol.~8, no.~6, pp. 620--628, 2014.

\bibitem{Su_2018_ECCV}
D.~Su, H.~Zhang, H.~Chen, J.~Yi, P.-Y. Chen, and Y.~Gao, ``Is robustness the
  cost of accuracy? {A} comprehensive study on the robustness of 18 deep image
  classification models,'' in \emph{European Conference on Computer Vision},
  2018.

\bibitem{YOLOV4}
A.~Bochkovskiy, C.~Wang, and H.~M. Liao, ``Yolov4: Optimal speed and accuracy
  of object detection,'' \emph{CoRR}, vol. abs/2004.10934, 2020.

\bibitem{Zhang2009}
E.~Zhang and Y.~Zhang, \emph{F-Measure}.\hskip 1em plus 0.5em minus 0.4em\relax
  Boston, MA: Springer US, 2009, pp. 1147--1147.

\bibitem{Goutte}
C.~Goutte and E.~Gaussier, ``A probabilistic interpretation of precision,
  recall and f-score, with implication for evaluation,'' in \emph{Proceedings
  of the 27th European Conference on Advances in Information Retrieval
  Research}, ser. ECIR'05.\hskip 1em plus 0.5em minus 0.4em\relax Berlin,
  Heidelberg: Springer-Verlag, 2005, p. 345–359.

\bibitem{LeNet}
Y.~LeCun, B.~Boser, J.~S. Denker, D.~Henderson, R.~E. Howard, W.~Hubbard, and
  L.~D. Jackel, ``Backpropagation applied to handwritten zip code
  recognition,'' \emph{Neural Computation}, vol.~1, no.~4, pp. 541--551, 1989.

\bibitem{Alex2012}
A.~Krizhevsky, I.~Sutskever, and G.~E. Hinton, ``Imagenet classification with
  deep convolutional neural networks,'' in \emph{Advances in Neural Information
  Processing Systems}, vol.~25, 2012.

\bibitem{Szegedy}
C.~Szegedy, V.~Vanhoucke, S.~Ioffe, J.~Shlens, and Z.~Wojna, ``Rethinking the
  inception architecture for computer vision,'' in \emph{IEEE Conference on
  Computer Vision and Pattern Recognition}, 2016, pp. 2818--2826.

\bibitem{EfficientNet}
M.~Tan and Q.~V. Le, ``Efficientnet: Rethinking model scaling for convolutional
  neural networks,'' in \emph{{PMLR} Proceedings of the 36th International
  Conference on Machine Learning}, vol.~97, 2019, pp. 6105--6114.

\bibitem{ViT}
A.~Dosovitskiy, L.~Beyer, A.~Kolesnikov, D.~Weissenborn, X.~Zhai,
  T.~Unterthiner, M.~Dehghani, M.~Minderer, G.~Heigold, S.~Gelly, J.~Uszkoreit,
  and N.~Houlsby, ``An image is worth 16x16 words: Transformers for image
  recognition at scale,'' in \emph{9th International Conference on Learning
  Representations}, 2021.

\bibitem{MLP_Mixer}
I.~O. Tolstikhin, N.~Houlsby, A.~Kolesnikov, L.~Beyer, X.~Zhai, T.~Unterthiner,
  J.~Yung, A.~Steiner, D.~Keysers, J.~Uszkoreit, M.~Lucic, and A.~Dosovitskiy,
  ``Mlp-mixer: An all-mlp architecture for vision,'' \emph{CoRR}, vol.
  abs/2105.01601, 2021.

\bibitem{Yarin2017}
R.~McAllister, Y.~Gal, A.~Kendall, M.~van~der Wilk, A.~Shah, R.~Cipolla, and
  A.~Weller, ``Concrete problems for autonomous vehicle safety: Advantages of
  bayesian deep learning,'' in \emph{Proceedings of the Twenty-Sixth
  International Joint Conference on Artificial Intelligence}, 2017, pp.
  4745--4753.

\bibitem{feng21}
D.~Feng, A.~Harakeh, S.~L. Waslander, and K.~Dietmayer, ``A review and
  comparative study on probabilistic object detection in autonomous driving,''
  \emph{IEEE Transactions on Intelligent Transportation Systems}, pp. 1--20,
  2021.

\bibitem{gledson_eccv}
G.~Melotti, C.~Premebida, J.~J. Bird, D.~R. Faria, and N.~Gonçalves,
  ``Probabilistic object classification using {CNN} {ML}-{MAP} layers,'' in
  \emph{Workshop on Perception for Autonomous Driving, European Conference on
  Computer Vision}, 2020.

\bibitem{Feng_Di}
D.~Feng, Z.~Wang, Y.~Zhou, L.~Rosenbaum, F.~Timm, K.~Dietmayer, M.~Tomizuka,
  and W.~Zhan, ``Labels are not perfect: Inferring spatial uncertainty in
  object detection,'' \emph{IEEE Transactions on Intelligent Transportation
  Systems}, pp. 1--14, 2021.

\bibitem{feng_eccv}
D.~Feng, L.~Rosenbaum, F.~Timm, and K.~Dietmayer, ``Labels are not perfect:
  Improving probabilistic object detection via label uncertainty,'' in
  \emph{Workshop on Perception for Autonomous Driving, European Conference on
  Computer Vision}, 2020.

\bibitem{patra}
R.~Patra, R.~Hebbalaguppe, T.~Dash, G.~Shroff, and L.~Vig, ``Calibrating deep
  neural networks using explicit regularisation and dynamic data pruning,'' in
  \emph{IEEE Winter Conference on Applications of Computer Vision (WACV)},
  2023, pp. 1541--1549.

\bibitem{Krishnan}
R.~Krishnan and O.~Tickoo, ``Improving model calibration with accuracy versus
  uncertainty optimization,'' in \emph{Advances in Neural Information
  Processing Systems}, vol.~33, 2020, pp. 18\,237--18\,248.

\bibitem{Mesquita}
D.~P.~P. {Mesquita}, L.~A. {Freitas}, J.~P.~P. {Gomes}, and C.~L.~C. {Mattos},
  ``{LS-SVR} as a bayesian {RBF} network,'' \emph{IEEE Transactions on Neural
  Networks and Learning Systems}, pp. 1--5, 2019.

\bibitem{Passalis}
N.~{Passalis}, M.~{Tzelepi}, and A.~{Tefas}, ``Probabilistic knowledge transfer
  for lightweight deep representation learning,'' \emph{IEEE Transactions on
  Neural Networks and Learning Systems}, pp. 1--10, 2020.

\bibitem{posch}
K.~{Posch} and J.~{Pilz}, ``Correlated parameters to accurately measure
  uncertainty in deep neural networks,'' \emph{IEEE Transactions on Neural
  Networks and Learning Systems}, vol.~32, no.~3, pp. 1037--1051, 2021.

\bibitem{Feng2019}
D.~{Feng}, L.~{Rosenbaum}, F.~{Timm}, and K.~{Dietmayer}, ``Leveraging
  heteroscedastic aleatoric uncertainties for robust real-time lidar {3D}
  object detection,'' in \emph{IEEE Intelligent Vehicles Symposium}, 2019, pp.
  1280--1287.

\bibitem{zouyu2019}
Y.~{Zou}, Z.~{Yu}, X.~{Liu}, B.~V. K.~V. {Kumar}, and J.~{Wang}, ``Confidence
  regularized self-training,'' in \emph{IEEE International Conference on
  Computer Vision}, 2019, pp. 5981--5990.

\bibitem{oncalibration}
C.~Guo, G.~Pleiss, Y.~Sun, and K.~Q. Weinberger, ``On calibration of modern
  neural networks,'' in \emph{Proceedings of the 34th International Conference
  on Machine Learning}, vol.~70, 2017, pp. 1321--1330.

\bibitem{GabrielPereyra}
\emph{Regularizing Neural Networks by Penalizing Confident Output
  Distributions}, ser. CoRR, arXiv: 1701.06548, 2017.

\bibitem{gal16}
Y.~Gal and Z.~Ghahramani, ``Dropout as a bayesian approximation: Representing
  model uncertainty in deep learning,'' in \emph{PMLR Proceedings of The 33rd
  International Conference on Machine Learning}, vol.~48, 2016, pp. 1050--1059.

\bibitem{durk}
D.~P. Kingma, T.~Salimans, and M.~Welling, ``Variational dropout and the local
  reparameterization trick,'' in \emph{Advances in Neural Information
  Processing Systems}, vol.~28.\hskip 1em plus 0.5em minus 0.4em\relax Curran
  Associates, Inc., 2015.

\bibitem{blundell}
C.~Blundell, J.~Cornebise, K.~Kavukcuoglu, and D.~Wierstra, ``Weight
  uncertainty in neural network,'' in \emph{PMLR Proceedings of the 32nd
  International Conference on Machine Learning}, vol.~37, 2015, pp. 1613--1622.

\bibitem{Kingma}
D.~Kingma and M.~Welling, ``Auto-encoding variational {B}ayes,'' in \emph{ICLR
  Proceedings 2nd International Conference on Learning Representations}, 2014.

\bibitem{ivalex}
A.~Graves, ``Practical variational inference for neural networks,'' in
  \emph{24th Advances in Neural Information Processing Systems}, vol.~24, 2011,
  pp. 2348--2356.

\bibitem{Jiacheng}
J.~Cheng and N.~Vasconcelos, ``Calibrating deep neural networks by pairwise
  constraints,'' in \emph{IEEE/CVF Conference on Computer Vision and Pattern
  Recognition (CVPR)}, 2022, pp. 13\,699--13\,708.

\bibitem{Frenkel}
L.~Frenkel and J.~Goldberger, ``Network calibration by temperature scaling
  based on the predicted confidence,'' in \emph{2022 30th European Signal
  Processing Conference (EUSIPCO)}, 2022, pp. 1586--1590.

\bibitem{isotonicregression}
\emph{Transforming Classifier Scores into Accurate Multiclass Probability
  Estimates}.\hskip 1em plus 0.5em minus 0.4em\relax Proceedings of the Eighth
  ACM SIGKDD International Conference on Knowledge Discovery and Data Mining,
  2002.

\bibitem{plattscaling}
J.~C. Platt, ``Probabilistic outputs for support vector machines and
  comparisons to regularized likelihood methods,'' in \emph{Advances Large
  Margin Classifiers}, 2000, pp. 61--74.

\bibitem{Weizh_pmlre}
W.~Li, G.~Dasarathy, and V.~Berisha, ``Regularization via structural label
  smoothing,'' in \emph{Proceedings of the Twenty Third International
  Conference on Artificial Intelligence and Statistics}, ser. Proceedings of
  Machine Learning Research, vol. 108.\hskip 1em plus 0.5em minus 0.4em\relax
  PMLR, 26--28 Aug 2020, pp. 1453--1463.

\bibitem{Rafael}
R.~M\"{u}ller, S.~Kornblith, and G.~E. Hinton, ``When does label smoothing
  help?'' in \emph{Advances in Neural Information Processing Systems},
  H.~Wallach, H.~Larochelle, A.~Beygelzimer, F.~d\textquotesingle
  Alch\'{e}-Buc, E.~Fox, and R.~Garnett, Eds., vol.~32.\hskip 1em plus 0.5em
  minus 0.4em\relax Curran Associates, Inc., 2019.

\bibitem{Kendall2017}
A.~Kendall and Y.~Gal, ``What uncertainties do we need in bayesian deep
  learning for computer vision?'' in \emph{Advances in Neural Information
  Processing Systems 30}, 2017, pp. 5574--5584.

\bibitem{Bishop}
C.~M. Bishop, \emph{Pattern Recognition and Machine Learning}.\hskip 1em plus
  0.5em minus 0.4em\relax Springer, 2006.

\bibitem{Conde_2022_BMVC}
P.~Conde and C.~Premebida, ``Adaptive-{TTA}: accuracy-consistent weighted test
  time augmentation method for the uncertainty calibration of deep learning
  classifiers,'' in \emph{33rd British Machine Vision Conference 2022, {BMVC}
  2022, London, UK, November 21-24, 2022}.\hskip 1em plus 0.5em minus
  0.4em\relax {BMVA} Press, 2022.

\bibitem{kristiadi2020}
A.~Kristiadi, M.~Hein, and P.~Hennig, ``Being bayesian, even just a bit, fixes
  overconfidence in relu networks,'' \emph{arXiv preprint arXiv:2002.10118},
  2020.

\bibitem{thulasidasan2019}
S.~Thulasidasan, G.~Chennupati, J.~A. Bilmes, T.~Bhattacharya, and S.~Michalak,
  ``On mixup training: Improved calibration and predictive uncertainty for deep
  neural networks,'' in \emph{Advances in Neural Information Processing Systems
  32}, 2019, pp. 13\,888--13\,899.

\bibitem{bulatov2015}
K.~B. Bulatov and D.~V. Polevoy, ``Reducing overconfidence in neural networks
  by dynamic variation of recognizer relevance,'' in \emph{Proceedings 29th
  European Conference on Modelling and Simulation}, 2015, pp. 488--491.

\bibitem{raudys2003}
{\v{S}}.~Raudys, R.~Somorjai, and R.~Baumgartner, ``Reducing the overconfidence
  of base classifiers when combining their decisions,'' in \emph{Multiple
  Classifier Systems}, 2003, pp. 65--73.

\bibitem{Feng2018}
D.~{Feng}, L.~{Rosenbaum}, and K.~{Dietmayer}, ``Towards safe autonomous
  driving: Capture uncertainty in the deep neural network for lidar {3D}
  vehicle detection,'' in \emph{IEEE 21st International Conference on
  Intelligent Transportation Systems}, 2018, pp. 3266--3273.

\bibitem{flipout2018}
Y.~Wen, P.~Vicol, J.~Ba, D.~Tran, and R.~Grosse, ``Flipout: Efficient
  pseudo-independent weight perturbations on mini-batches,'' in \emph{ICLR 6th
  International Conference on Learning Representations}, 2018.

\bibitem{Balaji}
B.~Lakshminarayanan, A.~Pritzel, and C.~Blundell, ``Simple and scalable
  predictive uncertainty estimation using deep ensembles,'' in \emph{Advances
  in Neural Information Processing Systems}, vol.~30, 2017, pp. 6402--6413.

\bibitem{lukasik20a}
M.~Lukasik, S.~Bhojanapalli, A.~Menon, and S.~Kumar, ``Does label smoothing
  mitigate label noise?'' in \emph{PMLR Proceedings of the 37th International
  Conference on Machine Learning}, vol. 119, 2020, pp. 6448--6458.

\bibitem{ConcreteB}
Y.~Gal, J.~Hron, and A.~Kendall, ``Concrete dropout,'' in \emph{31st Advances
  in Neural Information Processing Systems}, vol.~30, 2017.

\bibitem{Relaxed}
L.~Neumann, A.~Zisserman, and A.~Vedaldi, ``Relaxed softmax: Efficient
  confidence auto-calibration for safe pedestrian detection,'' in \emph{NIPS
  Workshop on Machine Learning for Intelligent Transportation System}, 2018.

\bibitem{corbi}
C.~Corbi\`{e}re, N.~THOME, A.~Bar-Hen, M.~Cord, and P.~P\'{e}rez, ``Addressing
  failure prediction by learning model confidence,'' in \emph{Advances in
  Neural Information Processing Systems}, vol.~32, 2019.

\bibitem{HendrycksG17}
D.~Hendrycks and K.~Gimpel, ``A baseline for detecting misclassified and
  out-of-distribution examples in neural networks,'' in \emph{5th International
  Conference on Learning Representations}, 2017.

\bibitem{Bingyuan}
B.~Liu, I.~B. Ayed, A.~Galdran, and J.~Dolz, ``The devil is in the margin:
  Margin-based label smoothing for network calibration,'' in \emph{IEEE/CVF
  Conference on Computer Vision and Pattern Recognition (CVPR)}, 2022, pp.
  80--88.

\bibitem{Youngbum}
Y.~Hur, E.~Yang, and S.~J. Hwang, ``A simple framework for robust
  out-of-distribution detection,'' \emph{IEEE Access}, vol.~10, pp.
  23\,086--23\,097, 2022.

\bibitem{Wang_2021_ICCV}
Y.~Wang, B.~Li, T.~Che, K.~Zhou, Z.~Liu, and D.~Li, ``Energy-based open-world
  uncertainty modeling for confidence calibration,'' in \emph{Proceedings of
  the IEEE/CVF International Conference on Computer Vision (ICCV)}, 2021, pp.
  9302--9311.

\bibitem{goodfellow2015}
I.~J. Goodfellow, J.~Shlens, and C.~Szegedy, ``Explaining and harnessing
  adversarial examples,'' \emph{CoRR, arXiv}, vol. 1412.6572, 2015.

\bibitem{ChrSzegedy}
C.~Szegedy, W.~Zaremba, I.~Sutskever, J.~Bruna, D.~Erhan, I.~Goodfellow, and
  R.~Fergus, ``Intriguing properties of neural networks,'' in
  \emph{International Conference on Learning Representations}, 2014.

\bibitem{LeaConf}
T.~DeVries and G.~W. Taylor, ``Learning confidence for out-of-distribution
  detection in neural networks,'' \emph{CoRR, arXiv :1802.04865}, 2018.

\bibitem{LiangLS18}
S.~Liang, Y.~Li, and R.~Srikant, ``Enhancing the reliability of
  out-of-distribution image detection in neural networks,'' in \emph{6th
  International Conference on Learning Representations}, 2018.

\bibitem{Gal2016}
Y.~Gal, ``Uncertainty in deep learning,'' Ph.D. dissertation, University of
  Cambridge, 2016.

\bibitem{GoodBengCour}
I.~J. Goodfellow, Y.~Bengio, and A.~Courville, \emph{Deep Learning}, ser.
  Adaptive Computation and Machine Learning.\hskip 1em plus 0.5em minus
  0.4em\relax MIT Press, 2016.

\bibitem{Molchanov}
D.~Molchanov, A.~Ashukha, and D.~Vetrov, ``Variational dropout sparsifies deep
  neural networks,'' in \emph{PMLR Proceedings of the 34th International
  Conference on Machine Learning}, vol.~70, 2017, pp. 2498--2507.

\bibitem{srivastava2014dropout}
N.~Srivastava, G.~Hinton, A.~Krizhevsky, I.~Sutskever, and R.~Salakhutdinov,
  ``Dropout: a simple way to prevent neural networks from overfitting,''
  \emph{The journal of machine learning research}, vol.~15, no.~1, pp.
  1929--1958, 2014.

\bibitem{Papoulis}
A.~Papoulis and U.~Pillai, \emph{\BIBforeignlanguage{English (US)}{Probability,
  random variables and stochastic processes}}, 4th~ed.\hskip 1em plus 0.5em
  minus 0.4em\relax McGraw-Hill, Nov. 2001.

\bibitem{AdditiveS}
D.~Valcarce, J.~Parapar, and {\'A}.~Barreiro, ``Additive smoothing for
  relevance-based language modelling of recommender systems,'' in
  \emph{Proceedings of the 4th Spanish Conference on Information Retrieval},
  2016.

\bibitem{SmoTec}
S.~F. Chen and J.~Goodman, ``An empirical study of smoothing techniques for
  language modeling,'' Harvard Computer Science Group Technical Report, Tech.
  Rep., 1998.

\bibitem{Lidstone}
G.~J. Lidstone, ``Note on the general case of the bayes-laplace formula for
  inductive or a posteriori probabilities,'' \emph{Transactions of the Faculty
  of Actuaries}, vol.~8, p. 182–192, 1920.

\bibitem{CIOU}
Z.~Zheng, P.~Wang, W.~Liu, J.~Li, R.~Ye, and D.~Ren, ``Distance-iou loss:
  Faster and better learning for bounding box regression,'' \emph{Proceedings
  of the AAAI Conference on Artificial Intelligence}, vol.~34, no.~07, pp.
  12\,993--13\,000, 2020.

\bibitem{cosinerate}
I.~Loshchilov and F.~Hutter, ``{SGDR:} stochastic gradient descent with warm
  restarts,'' in \emph{5th International Conference on Learning
  Representations}, 2017.

\bibitem{CBN}
Z.~Yao, Y.~Cao, S.~Zheng, G.~Huang, and S.~Lin, ``Cross-iteration batch
  normalization,'' in \emph{IEEE Conference on Computer Vision and Pattern
  Recognitio}, 2021.

\bibitem{yan2018SECOND}
Y.~Yan, Y.~Mao, and B.~Li, ``Second: Sparsely embedded convolutional
  detection,'' \emph{Sensors}, vol.~18, no.~10, p. 3337, 2018.

\bibitem{melotti_icarsc}
G.~{Melotti}, C.~{Premebida}, and N.~{Gonçalves}, ``Multimodal deep-learning
  for object recognition combining camera and {LIDAR} data,'' in \emph{IEEE
  International Conference on Autonomous Robot Systems and Competitions}, 2020,
  pp. 177--182.

\bibitem{Niculescu}
A.~Niculescu-Mizil and R.~Caruana, ``Predicting good probabilities with
  supervised learning,'' in \emph{Proceedings of the 22nd International
  Conference on Machine Learning}, 2005, pp. 625--632.

\end{thebibliography}

\end{document}